\documentclass[preprint,12pt]{elsarticle}

\usepackage{amssymb}
\usepackage{amsmath}

\usepackage{latexsym}
\usepackage{amssymb}
\usepackage{booktabs}
\usepackage{enumitem}
\usepackage{graphicx}
\usepackage{color}
\usepackage[small]{caption}
\usepackage{subcaption}
\usepackage{enumitem}

\usepackage{highlight}
\newcommand{\g}[1]{\gradientcelld{#1}{-0.4}{0}{0.4}{blue}{white}{red}{60}}
\usepackage{colortbl}
\usepackage{siunitx}

\usepackage{array}
\newcolumntype{L}[1]{>{\raggedright\arraybackslash}p{#1}}
\newcolumntype{C}[1]{>{\centering\arraybackslash}p{#1}}
\newcolumntype{R}[1]{>{\raggedleft\arraybackslash}p{#1}}

\usepackage{soul}

\journal{Information Sciences}

\begin{document}

\begin{frontmatter}



\title{Bootstrap Sampling Rate Greater than 1.0 May Improve Random Forest Performance} 

\author[label1]{Stanis{\l}aw Ka{\'z}mierczak\corref{correspondingAuthor}}
\ead{stanislaw.kazmierczak@pw.edu.pl}
\author[label1]{Jacek Ma{\'n}dziuk}
\ead{jacek.mandziuk@pw.edu.pl}
\affiliation[label1]{organization={Faculty of Mathematics and Information Science,
Warsaw University of Technology},
            addressline={Koszykowa 75},
            city={Warsaw},
            postcode={00-662},
            country={Poland}}

\cortext[correspondingAuthor]{Corresponding author}



\begin{abstract}
Random forests (RFs) utilize bootstrap sampling to generate individual training sets for each component tree by sampling with replacement, with the sample size typically equal to that of the original training set ($N$). Previous research indicates that drawing fewer than $N$ observations can also yield satisfactory results. The ratio of the number of observations in each bootstrap sample to the total number of training instances is referred to as the bootstrap rate (BR). Sampling more than $N$ observations (BR~$>$ 1.0) has been explored only to a limited extent and has generally been considered ineffective. In this paper, we revisit this setup using 36 diverse datasets, evaluating BR values ranging from 1.2 to 5.0. Contrary to previous findings, we show that higher BR values can lead to statistically significant improvements in classification accuracy compared to standard settings (BR~$\leq$~1.0). Furthermore, we analyze how BR affects the leaf structure of decision trees within the RF and investigate factors influencing the optimal BR. Our results indicate that the optimal BR is primarily determined by the characteristics of the data set rather than the RF hyperparameters. 
\end{abstract}



\end{frontmatter}


\section{Introduction}
\label{sec:Introduction}
Random forest (RF), introduced by \citet{breiman2001random}, is an ensemble of decision trees (DTs) that collectively make decisions using either majority or soft voting. RF reduces variance, sometimes at the cost of slightly increasing bias, by introducing two sources of randomness. The first one is the use of distinct random subsets of features when selecting the best split at each tree node. The second is training each tree on a subset of observations drawn with replacement from the original training set, i.e., a bootstrap sample.

In this study, we analyze the bootstrap rate (BR), an RF hyperparameter that controls the training process and consequently affects the model's performance. BR is defined as the ratio of the number of observations in each bootstrap sample to the total number of training instances. In the literature, this parameter is also referred to as the sample rate, subsample size, bootstrap size ratio, or bag size. In the original work, \citet{breiman2001random} used BR~$=$~1.0. However, lower values have also been successfully applied~\cite{adnan2014Improving,martinez2010Out,mansoori2023optimization,wozniacki2024novel}. When BR is low, each tree is trained on a smaller and more distinct subset of the data, which increases diversity among RF estimators. Naturally, the computational cost is reduced compared to BR~$=$~1.0. On the other hand, the trees may become too weak, as they are trained on a relatively smaller portion of the data. For BR~$=$~1.0, the expected fraction of unique observations in each bootstrap sample is 63.2\% of the dataset~\cite{efron1997improvements}, meaning that 36.8\% of observations are absent in each sample. When BR~$<$~1.0, the fraction of unique observations decreases even further.

It is not obvious what would happen if BR~$>$~1.0. 
On the one hand, a higher BR causes subsets to be less diverse, but on the other hand, it includes more unique observations (i.e., more information) in each sample. To our knowledge, \citet{martinez2010Out} are the only ones to analyze BR~$>$~1.0. However, they considered only BR~$=$~1.2 and concluded that this parameterization is generally ineffective. In our study, we not only analyze BR~$=$~1.2 (and lower) but also explore higher values of 2, 3, 4, and 5. Additionally, we extend the experimental setup to 18 RF configurations, whereas the reference paper seems to focus on a single configuration (though this is not clearly specified). Surprisingly, and in contrast to the findings of \citet{martinez2010Out}, we observe that BR~$>$~1.0 often yields better results than conventional BR values in the range $(0, 1]$.

The key contributions of this work can be summarized as follows:
\begin{itemize}
    \item To the best of our knowledge, this work is the first to investigate what the optimal BR value depends on;
    \item It is also the first study to suggest that exploring BR values greater than 1.0 is informative and often yields better results than the standard setting of BR~$\leq$~1.0;
    \item Our study reveals that employing small BR values ($\leq$~0.2), a setting that has received little attention in previous research, may enhance the predictive performance of RFs;
    \item We demonstrate that the optimal BR value is largely dataset-dependent and only partially influenced by the RF configuration;
    \item We empirically show that training time increases sublinearly with BR, indicating diminishing computational overhead for larger bootstrap ratios;
    \item We introduce neighborhood-based statistics whose strong correlation with the optimal BR provides empirical support for our theoretical analysis.
\end{itemize}

The rest of this paper is organized as follows. Section~\ref{sec:related_literature} reviews related literature. Section~\ref{sec:experiment_configuration} details the experimental configuration, including characteristics of the datasets, data preprocessing, tested hyperparameters, and the experimental design. Section~\ref{sec:results} reports the main results, focusing on the superiority of high vs.\ standard BR values, the distribution of optimal BR values, the shapes of BR curves, and time performance. Section~\ref{sec:understanding} provides a deeper analysis of the factors underlying the optimal BR, including proximity order, neighborhood structure, and the limitations of neighborhood-based analysis. Finally, Section~\ref{sec:conclusions} concludes the paper.

\section{Related Literature}
\label{sec:related_literature}
Hyperparameter tuning plays a critical role in the performance and efficiency of modern machine learning algorithms. In this section, we first review recent advances in hyperparameter optimization (HPO) across general machine learning contexts, including automation, efficiency, and privacy-aware settings. We then narrow the focus to RFs, where we discuss the most relevant hyperparameters and highlight the limited attention given to the BR.

\subsection{Hyperparameter Optimization}
Recent years have seen substantial progress in HPO across diverse ML contexts---from automated pipelines to reinforcement learning and privacy-sensitive applications. \citet{bischl2023hyperparameter} offer a comprehensive survey of HPO techniques, covering grid/random search, Bayesian methods, gradient- and population-based strategies, multi-objective/bandit-based and racing algorithms, as well as runtime optimizations and parallelization.

In the reinforcement learning domain, \citet{pmlr-v202-eimer23a} emphasize reproducibility and the influence of hyperparameter seed selection. In their empirical evaluation, the authors compare modern HPO tools to manual tuning across several RL algorithms and environments, finding that HPO methods often yield better performance with lower compute overhead. They therefore recommend AutoML-derived best practices---such as separating tuning and testing seeds and conducting principled HPO over broad search spaces.

For gradient-boosted trees, \citet{SOROKIN2023110604} present a model-aware hyperparameter tuning system that combines meta-learning with multi-fidelity optimization and automates the choice of the search space, thereby reducing the need for domain expertise in GBT tuning. \citet{INFSCI2019-Mantovani} demonstrate that a meta-learning recommender can predict when hyperparameter tuning will significantly improve performance (the study focused on SVMs across many datasets), which can substantially lower optimization cost. \citet{INFSCI2021-EstevezVelarde} propose a hierarchical AutoML framework (HML-Opt) that jointly optimizes pipeline structure and hyperparameters using grammatical evolution, reporting competitive benchmark performance. Complementary to these approaches, Auto-sklearn 2.0~\citep{feurer2022autosklearn} introduces PoSH (Portfolio Successive Halving) and warm-start meta-learning to allocate budget efficiently and yield strong results under strict time constraints.

A popular framework for implementing probabilistic models in HPO is Sequential Model-Based Optimization (SMBO)~\cite{10.1007/978-3-642-25566-3_40}, where the model is iteratively refined with an exploration--exploitation balance in mind. In each evaluation, exploration of new hyperparameter configurations is balanced with the exploitation of known high-performing areas. Well-known examples of SMBO methods are Gaussian Process Estimator (GPE)~\cite{snoek2012practicalbayesianoptimizationmachine} and the Tree Parzen Estimator (TPE)~\cite{10.5555/2986459.2986743}. Practical HPO tooling that implements these ideas (including dynamic search spaces and pruning) is exemplified by Optuna~\citep{akiba2019optuna}, which popularized the define-by-run API and efficient pruning/pruner implementations for scalable experiments. Production-ready SMBO packages such as SMAC3~\citep{lindauer2022smac3} provide multiple facades (SMAC4BB / SMAC4HPO / SMAC4MF), support multi-fidelity and algorithm-configuration use cases, and have been integrated into AutoML systems to make SMBO robust and practical at scale. In this line of research, \citet{Sieradzki2025EATPE} proposed the \mbox{EATAPE} method---an extension of TPE and its subsequent ATPE variant~\cite{arsenault2023learning}. For deep learning workloads, recent practical HPO algorithms such as PriorBand~\citep{mallik2023priorband} leverage expert priors and cheap proxy tasks to better align HPO with common DL experimental workflows, improving sample efficiency in realistic settings. Bandit-based early-stopping (Hyperband)~\citep{li2018hyperband} and hybrid approaches that combine model-based search with bandit-style pruning, e.g., BOHB~\citep{falkner2018bohb}, aim to offer strong anytime performance together with good final convergence. Large-scale HPO systems that address production constraints by automatic resource allocation and asynchronous scheduling---such as Hyper-Tune~\citep{li2022hypertune}---report substantial speedups over BOHB in real workloads. Surrogate-based benchmark suites such as YAHPO Gym~\citep{pfisterer2022yahpo} provide multi-fidelity, multi-objective testbeds that facilitate reproducible and scalable comparisons of HPO algorithms.

Another line of HPO research refers to hyperheuristic methods that rely on self-adaptation~\cite{Okulewiczetal2022,Okulewiczetal2020} or portfolio-based hybridizations of candidate algorithms aiming at selection and/or optimal parameterization of the best-suited method~\cite{Zakrzewski2025KES}. In a similar spirit, \citet{gecco2025,Zychowski2025KES,Zychowski2025Zakopane,Zajecka20245775} apply a portfolio of metaheuristics to solving TSP in an island-based optimization setup. Hybrid metaheuristic strategies are also gaining attention. \citet{CHM2025} propose a constrained hybrid metaheuristic (cHM) that dynamically combines multiple population-based algorithms to optimize the smoothing parameters of probabilistic neural networks, achieving consistent gains across diverse benchmark datasets. 

For multi-objective optimization beyond accuracy, \citet{dou2024hypertuner} present a cross-layer HPO framework that jointly optimizes model accuracy, training time, and energy consumption. Similarly, \citet{INFSCI2023-Viadinugroho} illustrate comparable trade-offs between accuracy and computational cost in sentiment analysis tasks. In related work on relational topic models, \citet{INFSCI2023-Terragni} emphasize that hyperparameter choices---such as topic priors or the number of topics---can lead to substantial differences in model behavior even under fixed structural assumptions.

When hyperparameter optimization must comply with differential privacy constraints, \citet{NEURIPS2023_82d7d58c} introduce DP-HyPO, an adaptive HPO framework designed to operate under $(\varepsilon,\delta)$-Differential Privacy (DP). The paper provides a formal privacy analysis and practical mechanisms for allocating and accounting for the privacy budget so that adaptive selection of candidates remains privacy-safe. Empirically, the authors demonstrate that DP-HyPO can substantially outperform naive (randomized) private tuning while respecting the same privacy budget.

Building on advances in automated end-to-end pipelines, \citet{filippou2023automl} demonstrate how structure learning can be seamlessly integrated with hyperparameter optimization within AutoML systems. Their framework jointly learns both model structure and hyperparameters, leading to more automated and robust model construction across diverse datasets.

A comparative evaluation in a healthcare context \citep{meaney2025comparison} conducts a benchmark of nine HPO strategies—including Bayesian TPE, Gaussian process models, evolutionary strategies, quasi-Monte Carlo, and simulated annealing—for tuning XGBoost models. The authors report that all HPO methods improved discrimination and calibration relative to default hyperparameters, although the relative advantages among methods vary depending on dataset characteristics such as size and signal-to-noise ratio.

\subsection{Random Forest Hyperparameter Tuning}
\citet{probst2019Hyperparameters}, in their survey on RF tuning, identify key hyperparameters commonly studied in RF optimization. The hyperparameter most extensively explored, the number of trees, was also analyzed by \citet{oshiro2012many}, \citet{scornet2017tuning}, and \citet{probst2018totune}. The optimization of the number of attributes considered during the splitting process was investigated by \citet{simon2009influence} and \citet{goldstein2011statistical}. Additionally, \citet{scornet2017tuning} and \citet{duroux2018impact} examined the influence of limiting tree depth.

The BR remains underexplored in the literature. \citet{probst2019Hyperparameters} consider its influence on RF performance to be minor, while also suggesting that tuning it can often yield improvements. \citet{duroux2018impact} argue that the inherent complexity of RFs makes conducting a thorough theoretical analysis challenging. As a result, many studies either omit bootstrapping altogether \cite{biau2008consistency,denil2013consistency,ishwaran2010consistency} or focus on simplified RF variants, such as median forests \cite{duroux2018impact,scornet2017tuning}.

The study most relevant to this work was conducted by~\citet{martinez2010Out}. To the best of our knowledge, this is the only work that analyzed BR~$>$~1.0, although the analysis is restricted to a single value of BR~$=$~1.2. The authors examine how RF performance depends on BR selection and identify four distinct types of BR curves that describe the relationship between BR and classification error across 30 datasets. However, the analysis is restricted to a single RF configuration and does not provide detailed insights into why the optimal BR varies significantly between datasets or why a specific curve shape corresponds to a particular dataset.

These observations highlight a critical gap in the literature: the role of the BR in RF performance has not been systematically examined beyond isolated scenarios. This study addresses this shortcoming by systematically evaluating a wide range of BR values across diverse datasets and RF configurations. The results reveal consistent patterns linking dataset characteristics to optimal BR values.

\section{Experiment Configuration}
\label{sec:experiment_configuration}
We conducted experiments on 36 diverse datasets, which underwent the following preprocessing steps: duplicate rows and rows corresponding to classes with only a single instance were removed. Columns with a single unique value were also dropped. Missing values in categorical features were replaced with a placeholder category, while missing values in numerical attributes were imputed using the column mean. Finally, one-hot encoding of categorical attributes was applied and numerical features were standardized. Table~\ref{tab:datasets} presents the characteristics of the datasets after the preprocessing.
Our experiments include all 30 datasets used by \citet{martinez2010Out} and six additional~ones.
\renewcommand{\arraystretch}{1.0}
\begin{table}
    \footnotesize
    \centering
    \caption {Dataset characteristics. The subsequent columns refer to the dataset name, the number of numerical and binary features, the number of observations, and the count of classes. The first 31 datasets come from the UCI Machine Learning Repository ~\citep{kelly2023UCI}. The next four are from ~\citet{breiman98arcing}, and the last one is from~\citet{breiman1984classification}.}
    \label{tab:datasets}
    \begin{tabular}{p{4.8cm}p{2.0cm}p{1.6cm}p{2.1cm}p{1.1cm}}
    \toprule
        Dataset & Numerical features & Binary features & Observations & Classes \\
    \midrule
        Abalone & 7 & 3 & 4172 & 23 \\
        Adult  & 6 & 85 & 48790 & 2 \\
        Arrhythmia & 194 & 64 & 420 & 12 \\
        Audiology (Standardized) & 0 & 89 & 171 & 18 \\
        Australian Credit Approval & 6 & 32 & 690 & 2 \\
        Balance Scale & 0 & 20 & 625 & 3 \\
        Breast Cancer Wisc. (Diag.) & 30 & 0 & 569 & 2 \\
        Breast Cancer Wisc. (Orig.) & 9 & 0 & 449 & 2 \\
        Congressional Voting Rec. & 0 & 48 & 342 & 2 \\
        Echocardiogram & 6 & 1 & 62 & 2 \\
        Ecoli & 5 & 1 & 336 & 8 \\
        German Credit Data & 6 & 53 & 1000 & 2 \\
        Glass Identification & 9 & 0 & 213 & 6 \\
        Heart & 7 & 13 & 270 & 2 \\
        Hepatitis & 6 & 27 & 148 & 2 \\
        Horse Colic & 7 & 140 & 368 & 2 \\
        Image Segmentation (Stat.) & 18 & 0 & 2086 & 7 \\
        Ionosphere & 32 & 1 & 350 & 2 \\
        Iris & 4 & 0 & 149 & 3 \\
        Labor Relations & 8 & 29 & 57 & 2 \\
        Liver Disorders & 5 & 0 & 341 & 2 \\
        Optical Recognition (Digits) & 61 & 0 & 1797 & 10 \\
        Parkinsons & 22 & 0 & 195 & 2 \\
        Pima Indians Diabetes & 8 & 0 & 768 & 2 \\
        Sonar, Mines vs. Rocks & 60 & 0 & 208 & 2 \\
        Soybean (Large) & 0 & 132 & 631 & 19 \\
        Tic-Tac-Toe Endgame & 0 & 27 & 958 & 2 \\
        Thyroid Disease & 5 & 0 & 215 & 3 \\
        Vehicle Silhouettes & 18 & 0 & 845 & 4 \\
        Vowel Recognition & 10 & 0 & 990 & 11 \\
        Wine & 13 & 0 & 178 & 3\\
        Ringnorm & 20 & 0 & 300 & 2 \\
        Threenorm & 20 & 0 & 300 & 2 \\
        Twonorm & 20 & 0 & 300 & 2 \\
        Waveform & 21 & 0 & 300 & 3 \\
        LED Display Domain & 0 & 24 & 200 & 10 \\
    \bottomrule
    \end{tabular}
\end{table}

The following hyperparameters (along with BR) are considered to be the most important for RF performance~\cite{probst2019Hyperparameters,scornet2017tuning,zhu2022optimization}:
\begin{itemize}
   \item 
number of trees ($nt$);
   \item 
parameters controlling the size of individual trees: maximum tree depth ($md$), the minimum number of instances required to split an internal node ($mn$), the minimum count of observations necessary to form a leaf node ($ml$);
   \item 
function measuring the quality of a split ($qs$);
   \item 
number of attributes considered when searching for the best split ($nf$).
\end{itemize}

As the base values for these hyperparameters, we adopted the defaults from the scikit-learn 1.1.3 Python package: $nt$~$=$~100, $md$~$=$~None (no depth limit), $qs$~$=$~"gini" (Gini impurity), $mn$~$=$~2, $ml$~$=$~1, $nf$~$=$~"sqrt" (square root of the number of features). We denote such a model as RF(base). Altogether, we tested RF(base) and 17 other configurations resulting from the following modifications of each single hyperparameter in RF(base):
\begin{itemize}
    \item RF(nt\_200), RF(nt\_500): number of trees equals 200 or 500, respectively;
    \item RF(md\_10), RF(md\_15), RF(md\_20), RF(md\_25): maximum depth of a tree equals 10, 15, 20, or 25, respectively;
    \item RF(qs\_ent): split quality is measured using Shannon entropy (information gain);
    \item RF(mn\_3), RF(mn\_4), RF(mn\_6), RF(mn\_8): minimum number of observations required to split an internal node is set to 3, 4, 6, or 8, respectively;
    \item RF(ml\_2), RF(ml\_3), RF(ml\_4), RF(ml\_5): minimum number of instances per leaf is 2, 3, 4, or 5, respectively;
    \item RF(nf\_log), RF(nf\_all): number of features considered at each split equals the logarithm with base 2 of the number of attributes or all features, respectively.
\end{itemize}

The following BR values were tested: 0.2, 0.4, 0.6, 0.8, 1.0, 1.2 (BR~=~1.2 was analyzed by \citet{martinez2010Out}), 2.0, 3.0, 4.0, and 5.0. For each configuration, 2-fold stratified cross-validation, repeated 200 times, was applied, yielding 400 results.

\section{Results}
\label{sec:results}
For each dataset, we identified the pair 
of (RF configuration, BR) that achieved the highest classification accuracy. For instance, $(RF(ml\_5), 0.2)$ for Abalone, $(RF(nt\_500), 1.2)$ for Wine, etc.  (hereafter referred to as the best or optimal RF configuration and/or BR). Table~\ref{tab:main_results} summarizes these results. Detailed outcomes, including the mean accuracy and standard deviation for each RF configuration and BR across all datasets, are provided in~\ref{appendix:Detailed_Results}.
\renewcommand{\arraystretch}{1.04}
\begin{table}
    \footnotesize
    \centering
    \caption {Classification results. The consecutive columns show the dataset name, the best RF configuration, the achieved accuracy, the optimal BR, and the $p$-value from the conducted $t$-test.}
    \label{tab:main_results}
    \begin{tabular}{p{4.8cm}p{2.6cm}p{1.9cm}p{0.9cm}p{1.4cm}}

    \toprule
        Dataset & Best model & Acc. [\%] & BR & $p$-value\\
    \midrule
        Abalone & RF(ml\_5) & 26.801 & 0.2 &  $< 10^{-6}$ \\
        Adult  & RF(ml\_5) & 86.484 & 4.0 & $< 10^{-6}$ \\
        Arrhythmia & RF(nf\_all) & 76.161 & 1.2 & 0.305022 \\
        Audiology (Standardized) & RF(mn\_8) & 75.338 & 5.0 & 0.013121 \\
        Australian Credit Approval & RF(nt\_500) & 87.225 & 0.6 & 0.132623 \\
        Balance Scale & RF(nt\_500) & 85.972 & 0.2 & $< 10^{-6}$ \\
        Breast Cancer Wisc. (Diag.) & RF(qs\_ent) & 95.898 & 5.0 & $< 10^{-6}$ \\
        Breast Cancer Wisc. (Orig.) & RF(nt\_500) & 95.506 & 0.4 & 0.001910 \\
        Congressional Voting Rec. & RF(mn\_8) & 94.795 & 2.0 & 0.029394 \\
        Echocardiogram & RF(ml\_5) & 73.113 & 2.0 & 0.035933 \\
        Ecoli & RF(nt\_500) & 85.835 & 0.6 & 0.000097 \\
        German Credit Data & RF(nt\_500) & 75.467 & 1.2 & 0.079419 \\
        Glass Identification & RF(qs\_ent) & 75.596 & 2.0 & 0.002702 \\
        Heart & RF(ml\_4) & 83.324 & 0.2 & 0.000004 \\
        Hepatitis & RF(nt\_500) & 84.726 & 0.4 & 0.000644 \\
        Horse Colic & RF(nt\_500) & 86.516 & 1.0 & 0.485986 \\
        Image Segmentation (Stat.) & RF(qs\_ent) & 97.133 & 5.0 & $< 10^{-6}$ \\
        Ionosphere & RF(nt\_500) & 93.254 & 1.2 & 0.192406 \\
        Iris & RF(mn\_4) & 95.232 & 0.4 & 0.105220 \\
        Labor Relations & RF(nt\_500) & 93.608 & 1.2 & 0.004194 \\
        Liver Disorders & RF(ml\_4) & 59.714 & 0.2 & $< 10^{-6}$ \\
        Optical Recognition (Digits) & RF(nt\_500) & 97.413 & 4.0 & $< 10^{-6}$ \\
        Parkinsons & RF(nt\_500) & 89.306 & 5.0 & $< 10^{-6}$ \\
        Pima Indians Diabetes & RF(nt\_500) & 76.344 & 0.2 & 0.000018 \\
        Sonar, Mines vs. Rocks & RF(qs\_ent) & 81.627 & 4.0 & $< 10^{-6}$ \\
        Soybean (Large) & RF(mn\_8) & 92.712 & 4.0 & $< 10^{-6}$ \\
        Tic-Tac-Toe Endgame & RF(nt\_500) & 97.264 & 5.0 & 0.021006 \\
        Thyroid Disease & RF(qs\_ent) & 95.840 & 1.2 & 0.059261 \\
        Vehicle Silhouettes & RF(nt\_500) & 74.583 & 5.0 & 0.001911 \\
        Vowel Recognition & RF(nt\_500) & 92.285 & 3.0 & $< 10^{-6}$ \\
        Wine & RF(nt\_500) & 97.809 & 1.2 & 0.072172 \\
        Ringnorm & RF(nt\_500) & 92.717 & 0.6 & $< 10^{-6}$ \\
        Threenorm & RF(nt\_500) & 80.050 & 0.4 & 0.000654 \\
        Twonorm & RF(nt\_500) & 96.002 & 0.2 & $< 10^{-6}$ \\
        Waveform & RF(nt\_500) & 86.165 & 0.2 & $< 10^{-6}$ \\
        LED Display Domain & RF(ml\_5) & 66.590 & 1.0 & $< 10^{-6}$ \\
    \bottomrule
    \end{tabular}
    \vspace{20pt}
\end{table}

\subsection{High vs. Standard BR Values} 
The main observation is that BR~$>$~1.0 constituted the best setup in 20 out of 36 datasets. To further compare standard BRs (BR~$\leq$~1.0, first group) with those greater than 1.0 (second group), we performed a paired $t$-test, testing the hypothesis that the first sample has a higher mean than the second one, on the results of the dataset winner (the best performing configuration) and all results from the other BR group. So, if the best classification accuracy was achieved by an RF configuration with BR~$\leq$~1.0, we compared its cross-validation results with all results related to configurations with BR~$>$~1.0, and vice versa. The last column of the table shows the maximum $p$-value among all \mbox{$t$-tests} for each dataset. We analyzed six significance levels: 0.1, 0.05, 0.01, 0.001, 0.0001, and 0.00001. Considering only conclusive results (i.e., $p$-values lower than the respective significance level), the difference in the number of datasets where the best model was associated with BR~$>$~1.0 versus those with BR~$\leq$~1.0 amounted to 5, 2, -2, -4, -2, and 0, respectively. This indicates that there is no clear advantage to either BR~$>$~1.0 or BR~$\leq$~1.0, suggesting that the optimal choice depends on dataset-specific characteristics.

\subsection{Distribution of Optimal BR Hyperparameter Values}
\label{subsec:br_distribution}
Among the 18 analyzed RF configurations, only seven achieved the highest classification accuracy in at least one dataset: RF(nt\_500) (20 datasets), RF(qs\_ent) (5~datasets), RF(ml\_5) (4 datasets), RF(mn\_8) (3 datasets), RF(ml\_4) (2 datasets), RF(mn\_4) (1~dataset), and RF(nf\_all) (1 dataset). The subsequent analysis focuses on these 7 configurations.

\begin{figure}
\centering
  \includegraphics[width=1.0\textwidth]{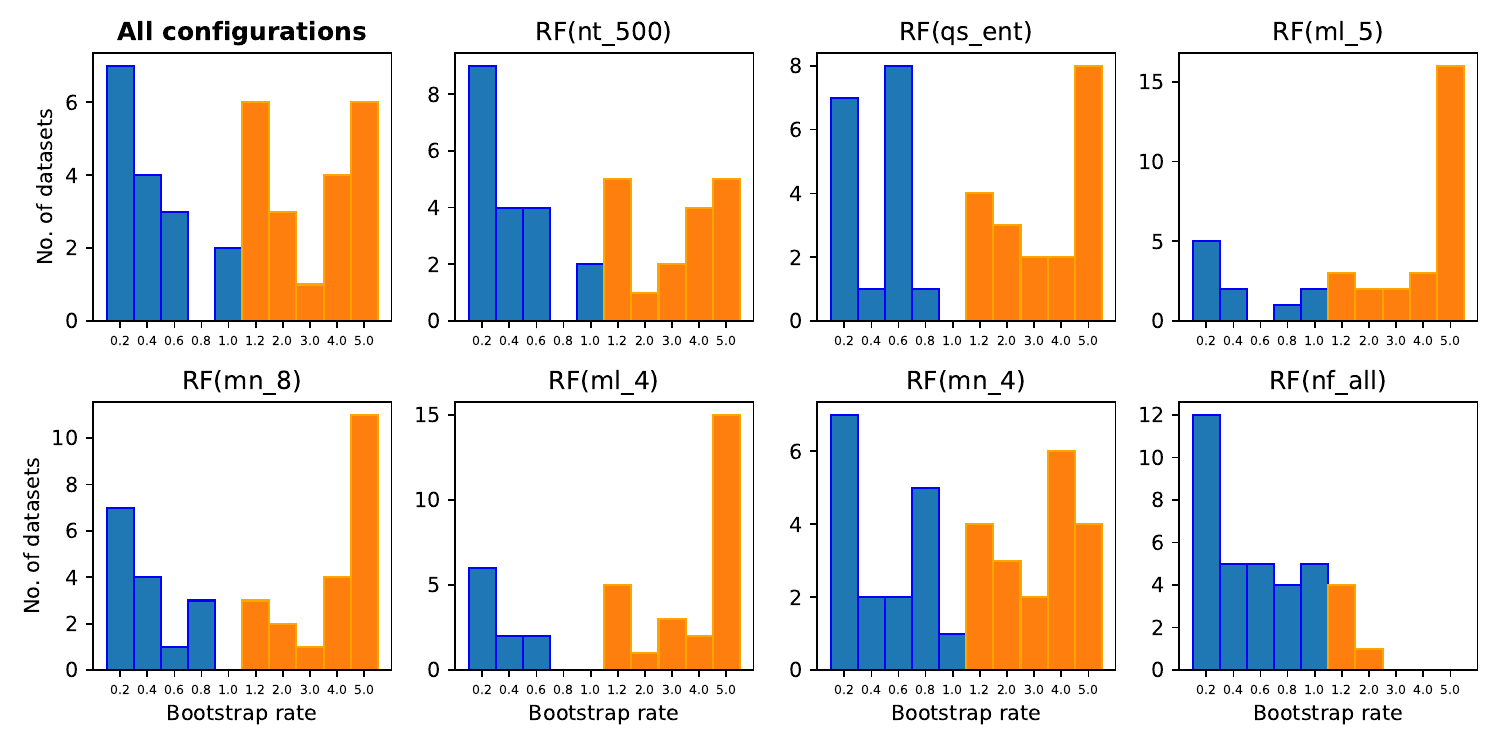}
\caption{Distribution of the winning BR across all RF configurations (top left) and among individual RF parameterizations.}
\label{fig:distribution}  
\end{figure}

Figure~\ref{fig:distribution} depicts the frequency of winning BRs, both globally and for each RF parameter setting. The histogram related to RF(nt\_500) is the closest to the global one, as this model achieved the best score in 20 out of 36 datasets. For RF(ml\_4) and RF(ml\_5), i.e., models that control the tree size more strictly, BR~$>$~1.0 constituted the best setup for as many as 26 datasets. The reason is that, in many cases, a low number of training instances combined with a relatively high minimum number of samples required to form a leaf led to underfitted trees. Thus, high BR helped mitigate this issue by enabling the construction of more complex models.

RF(nf\_all) exhibited distinct behavior compared to other models. The higher the BR, the less frequently it was optimal. The primary sources of diversity among individual trees are unique subsets of attributes considered for the best split in each node, along with the distinct bootstrap samples used during training. When all features are analyzed for node splitting, the first source of diversity is eliminated. Consequently, to preserve an overall level of diversity, RF(nf\_all) favored lower BRs, which produced more varied sample sets and reduced tree correlation. 

Analyzing the BR histograms, we observe that extreme BR values (0.2 and 5.0) most frequently constituted the best solutions, both overall and across all analyzed RF configurations, as indicated by the highest bars in each histogram corresponding to either 0.2 or 5.0. This suggests that the optimal BR may often be lower than 0.2 or higher than 5.0, implying that an even broader range of values should be considered when tuning RFs. Notably, BR~$=$~1.0, defined in the original formulation of the bootstrapping procedure and the value most frequently used in the literature, proved to be optimal for only two datasets overall. When analyzing individual RF configurations, for three of them, BR~$=$~1.0 did not yield the best performance in any dataset. Averaging across all seven hyperparameter settings, it was optimal for only 1.43 out of 36 datasets. Interestingly, adjacent to 1.0, the nonstandard BR~$=$~1.2 was, with the exception of RF(nf\_all), consistently superior, often substantially. 

\subsection{Analysis of BR Curve Shapes} 
\begin{figure}[b!]
\centering
  \includegraphics[width=1\textwidth]{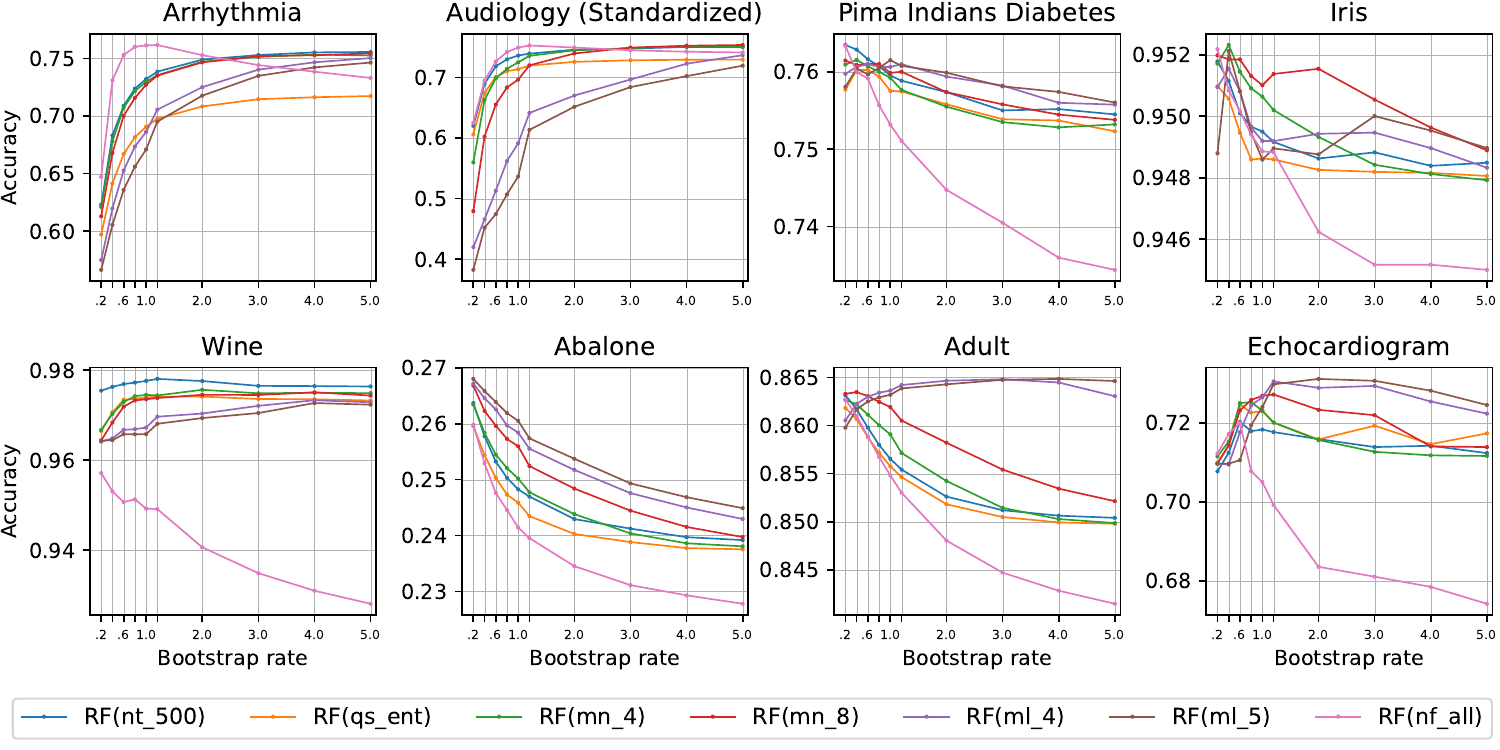}
\caption{BR curves for selected datasets.}
\label{fig:BR_characteristics_main}
\end{figure}

Figure~\ref{fig:BR_characteristics_main} illustrates the relationship between classification accuracy and BR for the analyzed RF configurations across a selected group of diverse datasets. The charts for the remaining datasets are provided in~\ref{appendix:BR_curves}.

The first notable observation is that RF(nf\_all) exhibits distinct behavior compared to the other models. In nearly all cases, it achieves optimal accuracy at a lower or equal BR relative to the other RF configurations. This finding aligns with the trends shown in Figure~\ref{fig:distribution} and the explanation provided in Section~\ref{subsec:br_distribution}. 
In most datasets, the maximum accuracy achieved by RF(nf\_all) is substantially lower than that of the other models, and its performance tends to decline sharply after reaching the optimum. However, there are several datasets where RF(nf\_all) performs well. It achieves the highest accuracy among all models on the Arrhythmia dataset and performs comparably to the best models on Audiology (Standardized), Tic-Tac-Toe Endgame, Pima Indians Diabetes, and Iris. 
The strong performance of RF(nf\_all) is likely related to dataset characteristics, particularly feature properties. For instance, the Arrhythmia dataset, where it performs best, has the highest number of features among all considered datasets.
However, the relationship appears to be more complex for the other four datasets. We hypothesize that further assessment of feature importance is needed to gain deeper insights. It is presumed that RF(nf\_all) may perform well on datasets with a high proportion of irrelevant or weakly informative features, as it can potentially avoid constructing trees that rely heavily on such features.

The BR curves for all RF configurations, except RF(nf\_all), generally exhibit similar characteristics. We identified three distinct categories that describe their behavior:
\begin{enumerate}[label=(\arabic*), leftmargin=*, align=left]
\item The first and most common pattern involves curves that increase up to at least BR~$=$~1.2, indicating that the optimal BR is not lower than 1.2. Beyond this point, the curves either continue to rise—typically more gradually—or reach a plateau, forming the first subpattern. Alternatively, they may oscillate or gradually decline, forming the second subpattern. The first subpattern is observed in the Arrhythmia, Audiology (Standardized), Parkinsons, Breast Cancer Wisc. (Diag.), Optical Recognition (Digits), Ionosphere, Image Segmentation (Stat.), Sonar, Mines vs. Rocks, Soybean (Large), Tic-Tac-Toe Endgame, Vowel, and Recognition datasets. The second subpattern is observed in the Wine, German Credit Data, Glass Identification, Labor Relations, Thyroid Disease, Vehicle Silhouettes, and Congressional Voting Rec. datasets.
\item In the second pattern, all curves either decrease from the very beginning (BR~$=$~0.2) or increase up to a BR within the range $[0.4, 1.0]$, and then decline. The overall shape of these curves may be relatively smooth, as observed in the Abalone, Balance Scale, Breast Cancer Wisc. (Orig.), Heart, Liver Disorders, Twonorm, Waveform, and LED Display Domain datasets, or it may exhibit some irregularities, as seen in the Iris and Pima Indians Diabetes datasets.
\item The third pattern is a combination of the first two. Curves associated with certain RF configurations, particularly RF(ml\_4) and RF(ml\_5), resemble the trends observed in the first pattern, while others follow those of the second. This mixed behavior is observed in the Adult, Australian Credit Approval, Ecoli, Hepatitis, Ringnorm, Threenorm, Horse Colic, and Echocardiogram datasets. For the last two datasets, additional irregularities in the BR curves can also be observed.
\end{enumerate}

The main observation from the above analysis is that BR curves for all RF configurations—except RF(nf\_all)—exhibit fairly consistent behavior. The first and second patterns, in which all curves follow similar trends, were observed in 28 out of 36 datasets. This suggests that the optimal BR is only weakly dependent on RF parameterization and is primarily determined by the dataset itself.

\subsection{Time Performance Across BR Values}
\begin{figure}
\centering
  \includegraphics[width=0.64\textwidth]{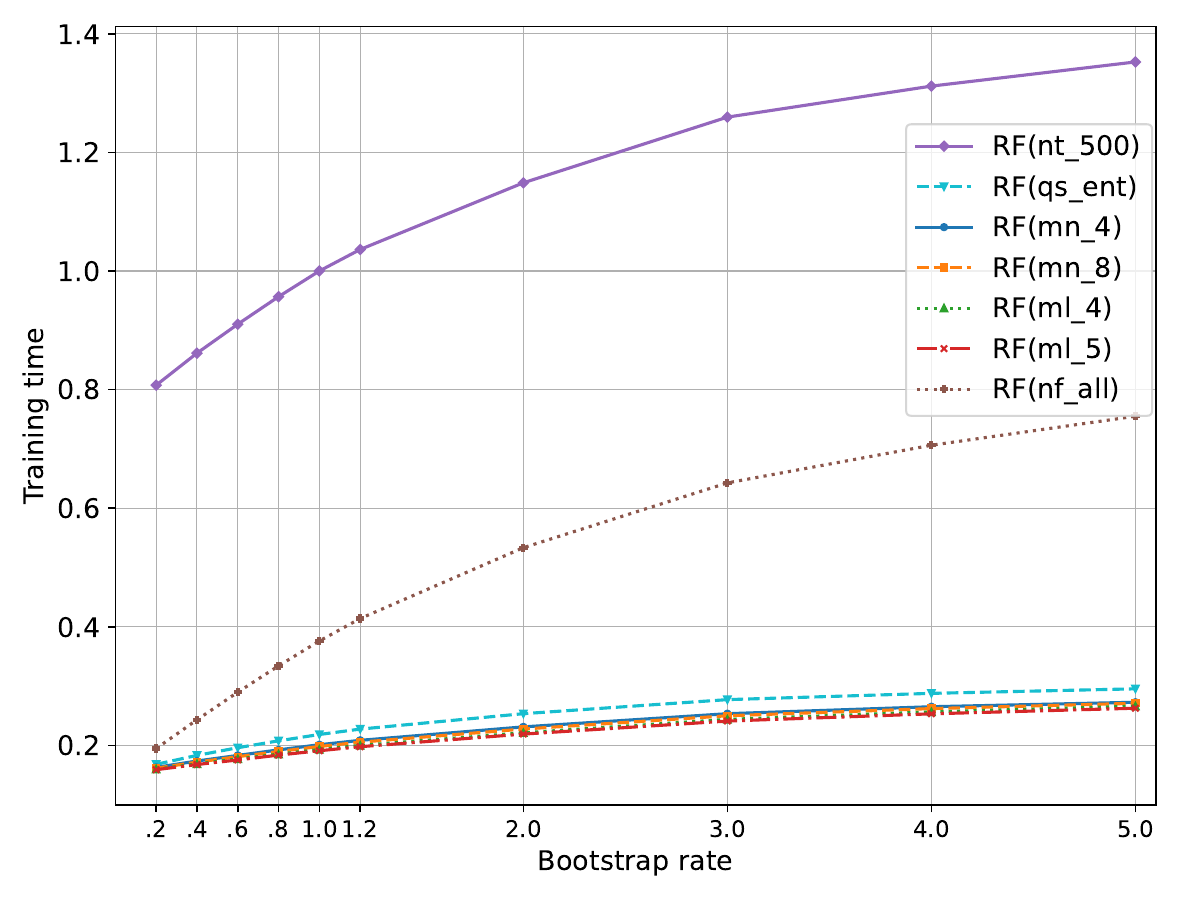}
\caption{The relationship between the training time of different RF configurations and the size of BR, averaged across all datasets. For each dataset, the times were normalized so that the training time of RF(nt\_500) with BR~$=$~1.0 equals 1.}
\label{fig:time_performance}
\end{figure}

Increasing the BR value yields a predictable trade-off: while larger bootstrap ratios can enhance predictive performance, they simultaneously increase computational cost due to the enlarged training subsets.
Figure~\ref{fig:time_performance} illustrates the relationship between training time and the BR value. Interestingly, the increase is sublinear across all analyzed configurations. We hypothesize that this is due to the number of possible split thresholds considered when splitting a node---as the bootstrap sample grows, the number of candidate splits increases at a slower rate. Furthermore, training time naturally scales with the number of trees in the RF, as demonstrated by RF(nt\_500), which is approximately five times slower than other configurations with 100 component trees. An exception is RF(nf\_all), which requires more time due to evaluating split thresholds across all features. Preliminary studies indicate that the shape of the training time curves with respect to BR values can vary significantly between datasets. A more in-depth analysis of this phenomenon may represent a valuable direction for future research.

\section{Understanding the Optimal BR}
\label{sec:understanding}
When exploring the reasons behind the significant differences in BR curves across datasets, we began by analyzing general dataset characteristics, such as the number of features (split into continuous and binary) and the number of training instances. We also engineered additional features by applying arithmetic operations to the aforementioned attributes to capture potential interactions. However, neither of these approaches improved our understanding of the phenomenon.

\begin{figure}
     \begin{subfigure}{0.495\textwidth}
         \centering
         \includegraphics[width=\textwidth]{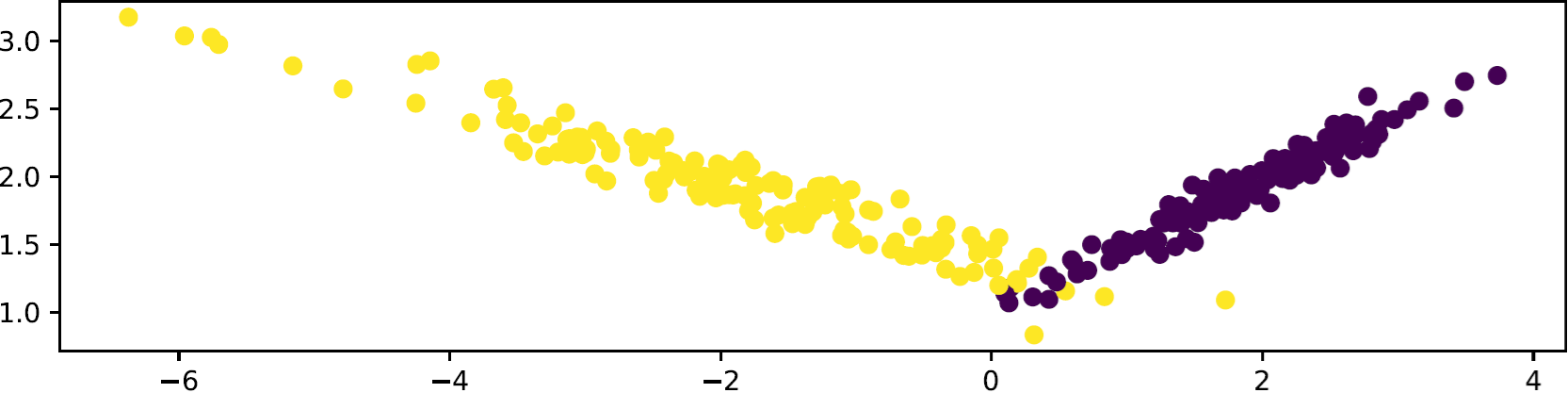}
         \caption{$class\_sep = 1.95$}
         \label{subfig:class_sep_1.95}
     \end{subfigure}
     \hfill
     \begin{subfigure}{0.495\textwidth}
         \centering
         \includegraphics[width=\textwidth]{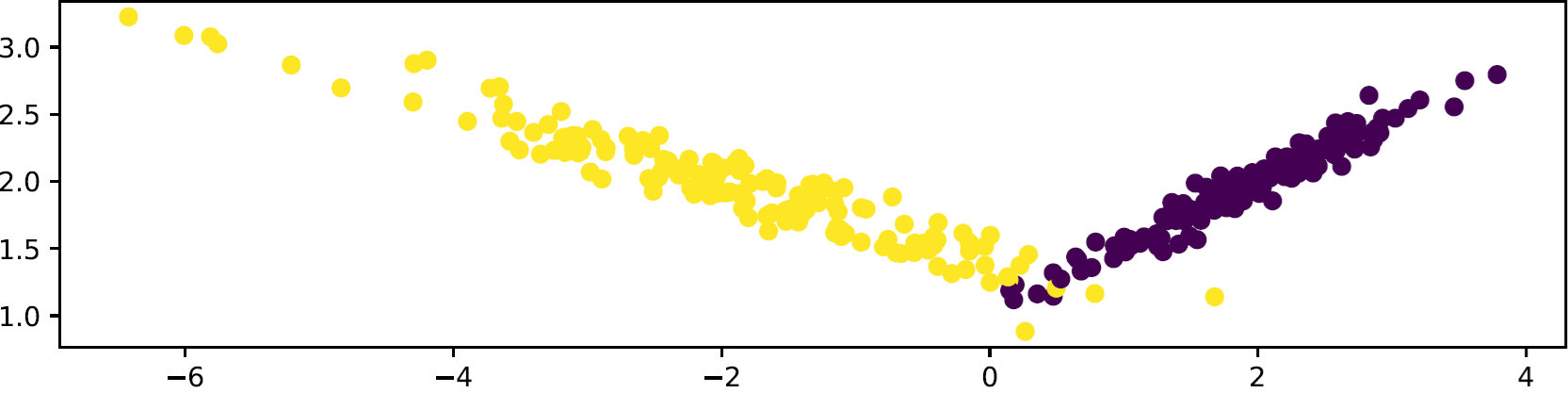}
         \caption{$class\_sep = 2.0$}
         \label{subfig:class_sep_2.0}
     \end{subfigure}
    \caption{An example illustrating how even small differences in the data can significantly affect the optimal BR value. Both figures (a) and (b) show synthetically generated data using scikit-learn's $make\_classification$ method with the following parameters: $n\_samples = 300$, $n\_features = 2$, $n\_classes = 2$, $n\_clusters\_per\_class = 1$, and $random\_state = 1$. The only difference between them is the value of the $class\_sep$ parameter, which controls class separation.  In (a), $class\_sep$ is set to 1.95, while in (b), it is set to 2.0. As a result of this slight difference, the optimal BR in (a) equals 5.0, while in (b), it amounts to 0.2. All other parameters of the $make\_classification$ method remain at their default values.}
    \label{fig:make_classification}
\end{figure}

Next, we adopted a more localized perspective and investigated a potential relationship between the BR curve and the number of clusters in the data. Unfortunately, this direction also proved inconclusive. In the meantime, we observed that even minor changes in the data could result in substantial differences in both the shape of the BR curve and its optimal value. Figure~\ref{fig:make_classification} illustrates such an example.
This insight led us to pursue an even more granular approach by analyzing the neighborhood structure of individual instances.

\subsection{Proximity Order and Its Impact on the Leaf Structure}
\label{subsec:proximity_order}
In brief, RF is composed of DTs that partition the feature space into decision regions, each defined by a path from the root to a leaf. In a single tree, the prediction for a sample falling into a particular leaf is based on the majority class of the training instances that reached that leaf. Intuitively, this implies that the predicted label depends on the local neighborhood of the sample—specifically, on the class distribution of nearby training instances. The same principle holds for RF, which aggregates predictions via voting across its constituent trees.
To better understand the structure of this neighborhood, we introduce the concept of a \emph{proximity order}---a strict partial order defined relative to a reference observation---which offers a more refined way of comparing instances in the feature space. By representing this order as a directed acyclic graph, we can analyze how BR affects the selection of training instances that form tree leaves. This, in turn, reveals how BR values shape the proximity and dispersion of training instances that influence predictions.

Let \( \mathcal{X} \subseteq \mathbb{R}^n \) denote a set of observations in an \( n \)-dimensional feature space, and let \( s \in \mathcal{X} \) be a reference observation. We define the proximity order \( \leq_s \) on \( \mathcal{X} \), relative to \( s \), as a strict partial order given by:
\begin{align}
a \leq_s b \iff \left( \forall i \in \{1, 2, \dots, n\}, \right. \nonumber \\
\left.\left( s_i \leq a_i \leq b_i \ \text{or} \ b_i \leq a_i \leq s_i \right)\right) \ \text{and} \ \left( \exists j \in \{1, 2, \dots, n\}, \right. \nonumber \\
\left.(a_j < b_j \ \text{if} \ s_j \leq a_j \leq b_j) \ \text{or} \ (b_j < a_j \ \text{if} \ b_j \leq a_j \leq s_j)\right)
\end{align}
where \( a = (a_1, \dots, a_n) \), \( b = (b_1, \dots, b_n) \), and \( s = (s_1, \dots, s_n) \).
This relation expresses that observation $a$ is \emph{closer to} $s$ than $b$ if and only if each feature value of $a$ lies between those of $s$ and $b$, and at least one feature differs between $a$ and $b$.

\begin{figure}
     \begin{subfigure}{0.483\textwidth}
         \centering
         \includegraphics[width=\textwidth]{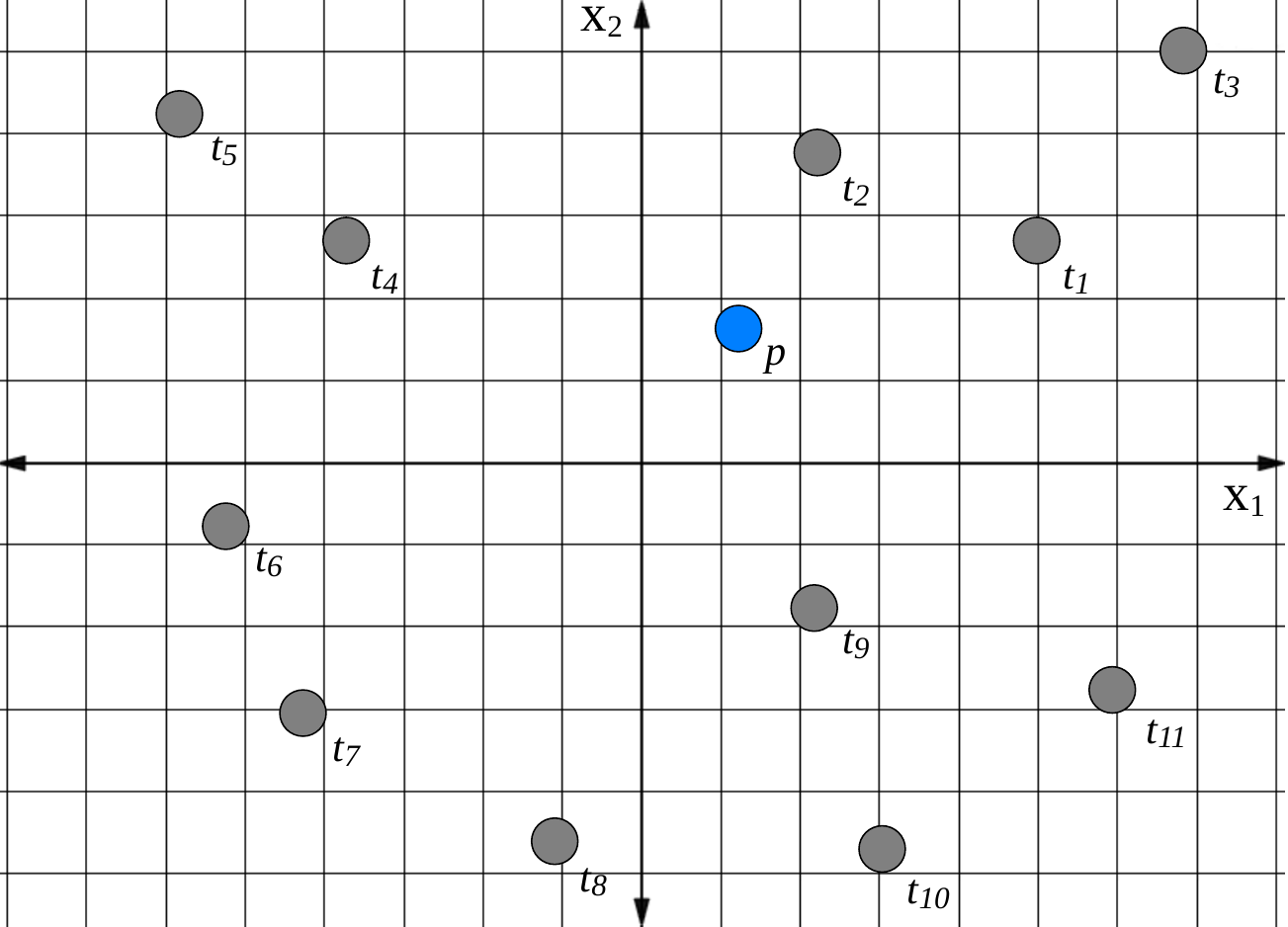}
         \caption{}
         \label{subfig:coordinates_example}
     \end{subfigure}
     \hfill
     \begin{subfigure}{0.483\textwidth}
         \centering
         \includegraphics[width=\textwidth]{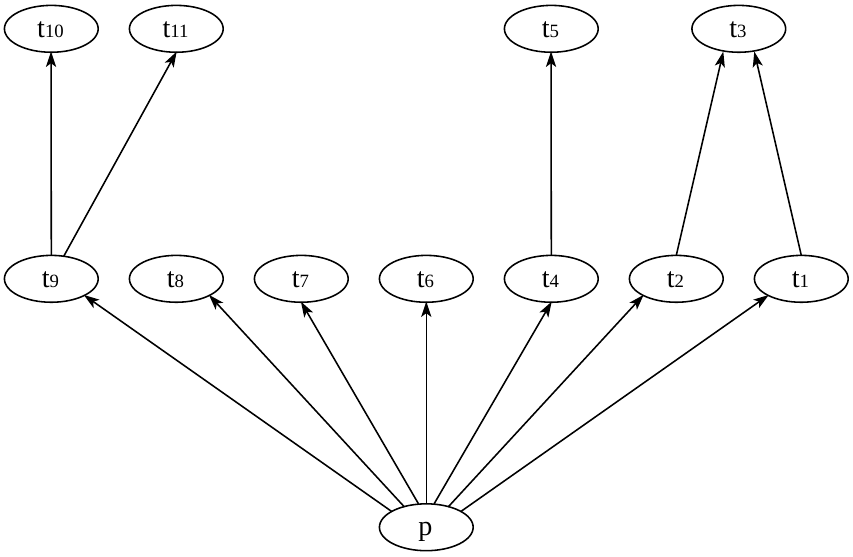}
         \caption{}
         \label{subfig:graph}
     \end{subfigure}
     \captionsetup{skip=12pt}
\caption{(a) presents an example of a training set ($t_1$--$t_{11}$) and a predicted observation $p$ in $\mathbb{R}^2$. (b) shows the Hasse diagram constructed from them.}
    \label{fig:partial_order}
\end{figure}

For the predicted observation $p$ and the training dataset $\mathcal{T}$, we construct a directed acyclic graph (Hasse diagram) $\mathcal{H}$ representing the relation $\leq_p$ for the observations in $\mathcal{T}$. Next, we add a vertex corresponding to the observation $p$ to the graph and connect it to all vertices that represent the minimal elements of $\mathcal{T}$ with respect to the relation $\leq_p$.
Figure~\ref{fig:partial_order} illustrates an example. The definition of the relation $\leq_p$ implies that if two vertices are connected by a $k$-edge path, and this is the only path connecting them, then the observations corresponding to the $k + 1$ vertices on this path may form a leaf in the DT. In particular, if $p$ is predicted based on a leaf formed from $k$ training observations, this leaf may contain observations that are up to $k$ edges away from $p$ in $\mathcal{H}$. This includes one edge from $p$ to the smallest element in the leaf, and $k - 1$ edges between the smallest and largest elements in the leaf.

\begin{figure}
     \begin{subfigure}{0.495\textwidth}
         \centering
         \includegraphics[width=\textwidth]{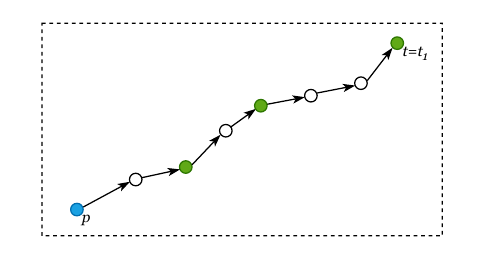}
         \caption{Lower BR}
     \end{subfigure}
     \hfill
     \begin{subfigure}{0.495\textwidth}
         \centering
         \includegraphics[width=\textwidth]{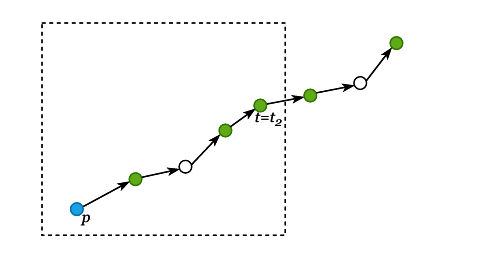}
         \caption{Higher BR}
     \end{subfigure}
     \captionsetup{skip=12pt}
\caption{ Panels (a) and (b) illustrate the path \(l\) from the predicted observation \(p\) to the training observation \(t\) that forms the leaf to which \(p\) is assigned during inference. Green and white circles denote training observations that are absent or present, respectively, in a given bootstrap sample. The dashed line encloses the observations forming the leaf into which \(p\) falls. A higher BR indicates that a larger fraction of training observations is included in each bootstrap sample. Assuming a constant number of observations forming the leaf along path \(l\) (three in this example), a higher BR implies that the prediction for \(p\) is influenced by training observations that are closer to \(p\), according to the \(\leq_p\) relation.}
    \label{fig:br_influence_on_leafs}
\end{figure}

The above analysis leads us to observe how BR influences bootstrap samples and, consequently, the structure of tree leaves. Let \( p \) denote the predicted instance. If a training observation \( t \in \mathcal{T} \) is located in the leaf to which \( p \) is assigned during inference, then all observations on the paths between \( p \) and \( t \) are also in that leaf. Let \( BS(l, br) \) be a random variable representing vertex numbers (measured as the path length from \( p \)) on the \( l \)-th path exiting from \( p \) in $\mathcal{H}$, given \( BR=br \), assuming the leaf is formed from a fixed number of observations. The probability of selecting any observation from $\mathcal{T}$ for the bootstrap sample is the same, and the leaf consists of those with the smallest numbers for each path \( l \). Thus, as \( br \) increases, the expected value \( E[BS(l, br)] \) decreases: the more observations included in the bootstrap sample, the closer (in terms of path length) the selected observations tend to be. Likewise, the density of selected observations increases with \( br \), leading to a decrease in the variance \( \text{Var}[BS(l, br)] \). In the limit, as \( br \to \infty \), the leaf would consist exclusively of the nearest neighbors of \( p \) in~\( \mathcal{H} \). 
Figure \ref{fig:br_influence_on_leafs} illustrates how the bootstrap ratio (BR) affects the composition of bootstrap samples and, consequently, the structure of the leaf to which the predicted observation \( p \) is assigned.

In conclusion, lower \( BR \) values cause the prediction to be influenced by more distant and dispersed observations, while higher values result in leaves formed by observations that are closer to the predicted instance and less dispersed. The optimal value of \( BR \) depends on both the dataset and RF configuration and is not constrained to the interval \( (0, 1] \), as demonstrated by our experiments.

\subsection{Empirical Analysis of Neighborhood Structure Using Manhattan Distance}
\label{subsec:empirical_analysis}
The analysis of neighborhood structure using the proximity order relation turned out to be difficult to apply in practice. Depending on the dataset, most of the considered observations had between a dozen and several dozen nearest neighbors. Many of them rarely or never appeared in the same leaf as the predicted instance due to their relatively large distance in the feature space. This effect was particularly visible when distance was measured using the Manhattan metric, which considers displacements along each feature (axis) independently. Since Manhattan distance aligns naturally with the structure of DTs---whose splits are axis-aligned and evaluate one feature at a time---we adopted it as the primary measure of proximity. For each dataset, continuous features were standardized, and binary features were mapped to -1 and 1.

Let $k\_l$ denote the number of observations for which exactly $l$ out of their $k$ nearest neighbors belong to the same class as the observation itself. Intuitively, high $k\_l$ values for small $l$ (relative to $k$) indicate class inhomogeneity in the local neighborhood and may signal a relatively large proportion of outliers. For each dataset, we computed the $k\_l$ statistics for $k \in \{1, 2, \dots, 10\}$ and $l \in \{0, \dots, k\}$. We then normalized these counts so that, for each fixed $k$, the sum $k\_0 + k\_1 + \dots + k\_k$ equaled 100, allowing for direct comparison across datasets of different sizes.

Across all datasets, we computed the Spearman correlation coefficient for each $k\_l$
 and for every RF configuration’s optimal BR (including the overall best). Table~\ref{tab:neighbors_correlation} reports the results for $k \in \{1, 2, \dots, 6\}$. Our first observation is that the overall best BR is consistently positively correlated with $k\_k$. The highest correlations correspond to $k = 1, 2, 3$, after which they gradually decrease. Second, for each $k\_l$ where $l \neq k$, the correlation is negative. For $k~\leq~5$, the lower $l$ is, the stronger the correlation becomes in absolute terms.

\renewcommand{\arraystretch}{1.09}
\begin{table}
    \footnotesize
    \centering
    \caption {Spearman rank-order correlation coefficient between $k\_l$ and the best BR---overall (second column) and respective RF configurations (columns 3--9).}

    \label{tab:neighbors_correlation}
    \begin{tabular}{p{0.608cm}|R{1.28cm}R{1.14cm}R{1.16cm}R{1.14cm}R{1.14cm}R{1.14cm}R{1.15cm}R{1.15cm}}
    \toprule
        $k\_l$ & Best RF & nt\_500 & qs\_ent\phantom{l} & ml\_5\phantom{m} & mn\_8\phantom{n} & ml\_4\phantom{m} & mn\_4\phantom{n} & nf\_all\phantom{n} \\
    \midrule
1\_0&\g{-0.311}&\g{-0.299}&\g{-0.345}&\g{-0.319}&\g{-0.312}&\g{-0.332}&\g{-0.387}&\g{-0.173}\\
1\_1&\g{0.311}&\g{0.299}&\g{0.345}&\g{0.319}&\g{0.312}&\g{0.332}&\g{0.387}&\g{0.173}\\
\midrule
2\_0&\g{-0.292}&\g{-0.252}&\g{-0.292}&\g{-0.263}&\g{-0.258}&\g{-0.280}&\g{-0.354}&\g{-0.156}\\
2\_1&\g{-0.252}&\g{-0.264}&\g{-0.298}&\g{-0.255}&\g{-0.241}&\g{-0.277}&\g{-0.317}&\g{-0.164}\\
2\_2&\g{0.330}&\g{0.301}&\g{0.347}&\g{0.320}&\g{0.320}&\g{0.332}&\g{0.379}&\g{0.163}\\
\midrule
3\_0&\g{-0.264}&\g{-0.258}&\g{-0.275}&\g{-0.242}&\g{-0.263}&\g{-0.256}&\g{-0.350}&\g{-0.139}\\
3\_1&\g{-0.250}&\g{-0.238}&\g{-0.311}&\g{-0.264}&\g{-0.250}&\g{-0.292}&\g{-0.331}&\g{-0.142}\\
3\_2&\g{-0.239}&\g{-0.213}&\g{-0.230}&\g{-0.164}&\g{-0.163}&\g{-0.192}&\g{-0.261}&\g{-0.183}\\
3\_3&\g{0.323}&\g{0.280}&\g{0.341}&\g{0.307}&\g{0.294}&\g{0.320}&\g{0.365}&\g{0.151}\\
\midrule
4\_0&\g{-0.292}&\g{-0.266}&\g{-0.278}&\g{-0.254}&\g{-0.268}&\g{-0.258}&\g{-0.351}&\g{-0.134}\\
4\_1&\g{-0.261}&\g{-0.233}&\g{-0.283}&\g{-0.255}&\g{-0.249}&\g{-0.274}&\g{-0.325}&\g{-0.159}\\
4\_2&\g{-0.213}&\g{-0.208}&\g{-0.264}&\g{-0.209}&\g{-0.179}&\g{-0.235}&\g{-0.280}&\g{-0.147}\\
4\_3&\g{-0.114}&\g{-0.116}&\g{-0.134}&\g{-0.031}&\g{-0.056}&\g{-0.067}&\g{-0.158}&\g{-0.090}\\
4\_4&\g{0.299}&\g{0.261}&\g{0.319}&\g{0.286}&\g{0.269}&\g{0.301}&\g{0.346}&\g{0.146}\\
\midrule
5\_0&\g{-0.238}&\g{-0.221}&\g{-0.218}&\g{-0.198}&\g{-0.185}&\g{-0.217}&\g{-0.285}&\g{-0.099}\\
5\_1&\g{-0.227}&\g{-0.178}&\g{-0.224}&\g{-0.183}&\g{-0.201}&\g{-0.205}&\g{-0.267}&\g{-0.034}\\
5\_2&\g{-0.223}&\g{-0.232}&\g{-0.298}&\g{-0.230}&\g{-0.220}&\g{-0.264}&\g{-0.321}&\g{-0.134}\\
5\_3&\g{-0.204}&\g{-0.148}&\g{-0.181}&\g{-0.114}&\g{-0.113}&\g{-0.143}&\g{-0.211}&\g{-0.127}\\
5\_4&\g{-0.084}&\g{-0.027}&\g{-0.035}&\g{0.096}&\g{0.056}&\g{0.055}&\g{-0.058}&\g{0.003}\\
5\_5&\g{0.302}&\g{0.244}&\g{0.301}&\g{0.269}&\g{0.245}&\g{0.285}&\g{0.318}&\g{0.125}\\
\midrule
6\_0&\g{-0.213}&\g{-0.186}&\g{-0.170}&\g{-0.165}&\g{-0.149}&\g{-0.182}&\g{-0.251}&\g{-0.080}\\
6\_1&\g{-0.156}&\g{-0.134}&\g{-0.183}&\g{-0.127}&\g{-0.165}&\g{-0.156}&\g{-0.223}&\g{0.031}\\
6\_2&\g{-0.234}&\g{-0.239}&\g{-0.292}&\g{-0.225}&\g{-0.214}&\g{-0.261}&\g{-0.328}&\g{-0.154}\\
6\_3&\g{-0.158}&\g{-0.128}&\g{-0.199}&\g{-0.167}&\g{-0.136}&\g{-0.198}&\g{-0.213}&\g{-0.015}\\
6\_4&\g{-0.109}&\g{-0.041}&\g{-0.062}&\g{0.039}&\g{0.039}&\g{0.008}&\g{-0.086}&\g{-0.083}\\
6\_5&\g{-0.013}&\g{0.030}&\g{-0.001}&\g{0.169}&\g{0.125}&\g{0.129}&\g{0.030}&\g{0.080}\\
6\_6&\g{0.265}&\g{0.204}&\g{0.261}&\g{0.220}&\g{0.190}&\g{0.235}&\g{0.260}&\g{0.080}\\
    \bottomrule
    \end{tabular}
\end{table}

These empirical patterns are consistent with the theoretical conclusions from Section~\ref{subsec:proximity_order}. When the nearest neighbors are of the same class as the predicted observation, a high BR increases the probability that this observation will fall into a leaf composed of those neighbors, thereby improving prediction accuracy. Conversely, the higher the $k\_0$, the more the model favors lower BR values, which lead to leaves formed from more distant and more dispersed observations. Hence, the chance of a correct prediction increases,
since nothing is known about neighbors beyond the $k$-th, while the closest ones are known to be of a different class. For $l \in \{1, \dots, k-1\}$, the correlation changes gradually, according to the degree of class agreement between the neighbors and the predicted observation. 

Optimal BRs associated with individual RF configurations exhibit properties similar to those of the best overall BR. For all $k$, the correlation between $k\_k$ and the optimal BR is positive, gradually decreasing after $2\_2$ (or $1\_1$ in the case of RF(nf\_all)). For $k \leq 5$, the remaining $k\_l$ values with $l \neq k$ generally show negative correlations, with some exceptions for $5\_4$. The relationship between $k\_0$ and $k\_k$ is generally upward, but, unlike for the best overall BR, it is non-monotonic. RF(nf\_all) exhibits a similar pattern, but the range of $k\_l$ values is substantially narrower, and more irregularities occur.

For both the overall best BR and the individual RF configurations, as $k$ increases beyond $k = 6$, the generally ascending correlation trend from $k\_0$ to $k\_k$ is maintained. However, correlation values become increasingly attenuated---their absolute magnitudes decrease, and irregularities emerge that are absent for smaller~$k$. This suggests that more distant neighbors, as defined by the Manhattan metric, occur less frequently in the leaves containing the predicted observation.

\subsection{Impact of Closer and More Distant Neighborhoods on Optimal BR}
\label{subsec:impact_closer_distant}
In Section~\ref{subsec:proximity_order} we observed that, in general, larger BR values result in the training observations within the leaf containing the predicted observation being, on average, closer to it. When the similarity between the training observations and the predicted instance is not substantially affected by the number of nearest neighbors considered, BR has little influence on the prediction accuracy for that instance. For example, if the broader neighborhood belongs to the same class as the predicted instance, both low and high BR values will produce leaves composed predominantly of observations from that class. Conversely, BR can have a substantial impact on model performance when the characteristics of the closer and more distant neighborhoods differ. The analysis of the correlation between $k\_l$ and the optimal BR value presented in Section~\ref{subsec:empirical_analysis} considers only the $k$ nearest neighbors. To incorporate both the closer and more distant neighborhoods, for each pair of distinct $k\_l$ statistics we generated derived statistics in the form of their ratio (both orders), product, sum, and difference. For ratio-based statistics, if both values were equal to zero, the ratio was set to~1; if only the denominator was zero, it was replaced with $10^{-6}$. Table~\ref{tab:derived_statistics} presents the ten derived statistics most strongly correlated with the overall optimal~BR.

\begin{table}
\centering
\caption{Top ten derived statistics with the highest positive correlation to the overall optimal BR. The ten most strongly negatively correlated statistics, not shown, are inverse ratios of these and exhibit nearly identical absolute correlation values.}
\label{tab:derived_statistics}
\begin{tabular}{r r r}
\toprule
No. & Statistic & Correlation \\
\midrule
1  & $9\_2 / 2\_0$   & 0.618 \\
2  & $10\_2 / 3\_0$  & 0.610 \\
3  & $9\_2 / 3\_0$   & 0.602 \\
4  & $10\_2 / 2\_0$  & 0.581 \\
5  & $10\_2 / 4\_0$  & 0.555 \\
6  & $9\_2 / 4\_1$   & 0.539 \\
7  & $8\_2 / 2\_0$   & 0.518 \\
8  & $9\_2 / 4\_0$   & 0.506 \\
9  & $9\_2 / 1\_0$   & 0.495 \\
10 & $10\_2 / 1\_0$  & 0.493 \\
\bottomrule
\end{tabular}
\end{table}

All ten of the derived statistics most strongly correlated with the optimal BR take the form of a ratio of two distinct $k\_l$ statistics. Moreover, all numerators and all denominators are similar in terms of their $k$ and $l$ values, and thus capture comparable structural aspects of the neighborhood. More specifically, each numerator refers to a broad neighborhood (8--10 neighbors) in which exactly two observations belong to the same class as the predicted instance. Each denominator represents a closer neighborhood (1--4 neighbors) containing either no neighbors of the same class (nine denominators) or exactly one such neighbor (one denominator). The corresponding correlation coefficient values exceed that of $2\_2$, the most highly correlated base statistic (Table~\ref{tab:neighbors_correlation}), by amounts ranging from 0.163 to 0.288.

The value of a ratio statistic increases either when its numerator increases or when its denominator decreases. The positive correlation shown in Table~\ref{tab:derived_statistics} indicates that the optimal BR value increases as the given statistic increases. Therefore, an increase in the numerator or a decrease in the denominator is conducive to a higher optimal BR, which is consistent with the theoretical considerations discussed in Section~\ref{subsec:proximity_order}. Indeed, a low denominator value means that few predicted observations have all of their closest neighbors belonging to a different class. Consequently, most observations have at least one (and often more than one) neighbor of the same class in their immediate vicinity, which supports higher BR values. A high numerator value, in turn, indicates that many observations have exactly two neighbors of the same class within their broader (both near and distant) neighborhood. Combined with the fact that some of these neighbors are located in the closer vicinity—as suggested by a low denominator value—one can infer that a relatively large number of observations have at most one (and often none) neighbor of the same class in their more distant neighborhood. Putting together both the above interpretations of the numerator and the denominator, it can be concluded that as the former increases and the latter decreases, the density of observations belonging to the same class as the predicted observation increases in its closer neighborhood and decreases in the more distant one. This favors higher BR values, which is consistent with the conclusions from Section~\ref{subsec:proximity_order}. In particular, larger BR causes the leaves to contain observations located closer to the predicted one.

\subsection{Limitations of Neighborhood-Based Analysis}
In this section, we examine dataset properties that determine the optimal BR. As discussed in Section~\ref{subsec:proximity_order}, the leaf to which a predicted observation is assigned is formed from neighboring training observations. Accordingly, in Sections~\ref{subsec:empirical_analysis} and~\ref{subsec:impact_closer_distant}, we introduced metrics capturing the neighborhood structure. The Spearman correlation coefficients of the statistics most strongly associated with the optimal BR, shown in Table~\ref{tab:derived_statistics}, exceed 0.6, with corresponding $p$-values below 0.001. It is very likely that neighborhood-based statistics could be developed that are even more strongly correlated with the optimal BR. Promising directions include a more detailed characterization of the neighborhood by incorporating additional base statistics $k\_l$. Another avenue is to investigate different feature normalization methods, which do not affect DT performance but fundamentally influence distance measurements that define the neighborhood.

Studying the impact of BR on RF quality using neighborhood analysis (i.e., a kNN-based approach) has certain limitations, stemming from the functional differences between DTs (built \emph{top-down}), and kNN (operating locally). Consequently:
\begin{itemize}
    \item When measuring distance, kNN considers all features, whereas a DT considers less important features less frequently or ignores them altogether.
    \item Some decision boundaries forming the leaves are determined at higher levels of the tree based on the distribution of features across the entire or a large portion of the training set. In contrast, kNN decision boundaries consider only the nearest observations.
    \item Decision boundaries in a tree are hyperrectangles formed through a sequence of splits in feature space, while kNN boundaries are irregular and often curved.
    \item kNN is highly sensitive to local anomalies (e.g., a single noisy point near a boundary). DTs, constrained by a minimum number of observations per leaf, aggregate data in leaves and can thus ignore individual anomalies.
\end{itemize}
The analysis is further complicated by the fact that the model considered is an RF, an ensemble of DTs that collectively make decisions. Consequently, the decisions of individual models are \emph{masked} by the majority vote. For these reasons, the neighborhood structure analysis provides only a partial answer regarding the optimality of BR, and a complete explanation without prior model training remains a very challenging task.

\section{Conclusions}
\label{sec:conclusions}
In this paper, we analyze the selection of the BR hyperparameter in RF. To the best of our knowledge, this is the first work to examine the factors that determine the optimal BR value. We demonstrate that the optimal BR often exceeds 1.0, going beyond the typically considered $(0, 1]$ range. We also show that the optimal BR value is largely independent of the other RF hyperparameters. In fact, most RF configurations exhibit relatively high correlations across the BR curve, suggesting that the optimal BR value is largely a dataset-specific property.

Our main conclusion is that BR~$>$~1.0 often yields superior results and deserves consideration. This contradicts the findings of the baseline study~\cite{martinez2010Out}. We attribute this discrepancy to two main factors. First, in~\cite{martinez2010Out}, the analysis is limited to BR~$=$~1.2 and does not explore higher BR values. Second, the results presented in~\cite{martinez2010Out} are based on a single RF configuration only, and without providing any details regarding the RF hyperparameters, except that the ensemble consists of 200 unpruned CART~\cite{breiman1984classification} trees.

We theoretically demonstrated how BR influences the structure of DT leaves. Low BR values effectively reduce the density of training samples, leading to leaves formed from observations that lie farther apart in the feature space. Conversely, higher BR values favor closer and less dispersed observations. As a result, the optimal BR value strongly depends on the local class structure. This was empirically shown by measuring the correlation of different $k\_l$ statistics, as well as more complex ones, derived statistics based on them, with the optimal BR.

An interesting avenue for future research is the development of models that predict the optimal BR value directly from dataset characteristics. Our results suggest that the optimal BR appears to be related to the local structure of the data, which can be captured through neighborhood-based statistics. Building a predictive model for BR would not only provide deeper insights into its dependence on dataset properties but would also offer a practical tool for selecting BR without exhaustive hyperparameter tuning.



Finally, we examined three widely used ML libraries: scikit-learn, Weka, and H2O.ai. In all of them, BR hyperparameter values greater than 1.0 are disabled in their RF implementations. Based on our findings, we recommend that the developers of ML libraries consider enabling this option.

\clearpage
\appendix
\section{Detailed Results}
\setcounter{table}{0}
\label{appendix:Detailed_Results}

\begin{table}[h!]
\centering
\caption{Classification accuracy (mean ± standard deviation) for the Abalone dataset.}
{\fontsize{5pt}{7pt}\selectfont
\begin{tabular}{p{0.18cm}|p{1.5153cm}p{1.5153cm}p{1.5153cm}p{1.5153cm}p{1.5153cm}p{1.5153cm}p{1.5153cm}}
\toprule
 BR &     RF(nt\_500) &     RF(qs\_ent) &       RF(mn\_4) &       RF(mn\_8) &       RF(ml\_4) &       RF(ml\_5) &     RF(nf\_all) \\
\midrule
0.2 & 26.368 ± 0.719 & 25.959 ± 0.753 & 26.345 ± 0.759 & 26.682 ± 0.766 & 26.718 ± 0.711 & 26.801 ± 0.703 & 25.984 ± 0.799 \\
0.4 & 25.777 ± 0.744 & 25.439 ± 0.769 & 25.835 ± 0.764 & 26.229 ± 0.754 & 26.457 ± 0.746 & 26.584 ± 0.714 & 25.290 ± 0.766 \\
0.6 & 25.323 ± 0.741 & 25.032 ± 0.750 & 25.447 ± 0.748 & 25.963 ± 0.774 & 26.256 ± 0.750 & 26.388 ± 0.740 & 24.761 ± 0.781 \\
0.8 & 25.031 ± 0.746 & 24.732 ± 0.780 & 25.206 ± 0.767 & 25.729 ± 0.747 & 25.970 ± 0.775 & 26.190 ± 0.803 & 24.458 ± 0.777 \\
1.0 & 24.825 ± 0.750 & 24.588 ± 0.719 & 25.017 ± 0.762 & 25.595 ± 0.762 & 25.840 ± 0.775 & 26.050 ± 0.730 & 24.142 ± 0.731 \\
1.2 & 24.696 ± 0.726 & 24.346 ± 0.720 & 24.773 ± 0.721 & 25.243 ± 0.761 & 25.553 ± 0.735 & 25.738 ± 0.770 & 23.953 ± 0.724 \\
2.0 & 24.294 ± 0.752 & 24.027 ± 0.726 & 24.383 ± 0.751 & 24.840 ± 0.768 & 25.173 ± 0.771 & 25.373 ± 0.797 & 23.452 ± 0.776 \\
3.0 & 24.123 ± 0.747 & 23.883 ± 0.739 & 24.038 ± 0.776 & 24.445 ± 0.773 & 24.758 ± 0.739 & 24.932 ± 0.753 & 23.112 ± 0.814 \\
4.0 & 23.970 ± 0.710 & 23.775 ± 0.714 & 23.861 ± 0.740 & 24.154 ± 0.738 & 24.502 ± 0.772 & 24.686 ± 0.753 & 22.930 ± 0.787 \\
5.0 & 23.918 ± 0.725 & 23.753 ± 0.760 & 23.806 ± 0.723 & 23.972 ± 0.733 & 24.295 ± 0.713 & 24.489 ± 0.800 & 22.779 ± 0.835 \\
\bottomrule
\end{tabular}
}
\end{table}

\begin{table}[h!]
\centering
\caption{Results for the Adult dataset.}
{\fontsize{5pt}{7pt}\selectfont
\begin{tabular}{p{0.18cm}|p{1.5153cm}p{1.5153cm}p{1.5153cm}p{1.5153cm}p{1.5153cm}p{1.5153cm}p{1.5153cm}}
\toprule
 BR &     RF(nt\_500) &     RF(qs\_ent) &       RF(mn\_4) &       RF(mn\_8) &       RF(ml\_4) &       RF(ml\_5) &     RF(nf\_all) \\
\midrule
0.2 & 86.315 ± 0.151 & 86.182 ± 0.152 & 86.266 ± 0.146 & 86.329 ± 0.152 & 86.053 ± 0.160 & 85.977 ± 0.157 & 86.270 ± 0.150 \\
0.4 & 86.168 ± 0.151 & 86.068 ± 0.154 & 86.220 ± 0.144 & 86.347 ± 0.146 & 86.224 ± 0.155 & 86.167 ± 0.157 & 86.098 ± 0.154 \\
0.6 & 85.977 ± 0.153 & 85.879 ± 0.157 & 86.112 ± 0.152 & 86.305 ± 0.146 & 86.300 ± 0.154 & 86.250 ± 0.152 & 85.886 ± 0.160 \\
0.8 & 85.798 ± 0.159 & 85.717 ± 0.165 & 86.002 ± 0.158 & 86.245 ± 0.157 & 86.341 ± 0.154 & 86.292 ± 0.153 & 85.673 ± 0.168 \\
1.0 & 85.654 ± 0.156 & 85.576 ± 0.157 & 85.908 ± 0.150 & 86.192 ± 0.150 & 86.363 ± 0.151 & 86.319 ± 0.155 & 85.482 ± 0.171 \\
1.2 & 85.542 ± 0.157 & 85.464 ± 0.155 & 85.715 ± 0.158 & 86.052 ± 0.152 & 86.420 ± 0.152 & 86.385 ± 0.155 & 85.302 ± 0.171 \\
2.0 & 85.261 ± 0.157 & 85.182 ± 0.158 & 85.424 ± 0.156 & 85.822 ± 0.153 & 86.465 ± 0.153 & 86.428 ± 0.157 & 84.805 ± 0.174 \\
3.0 & 85.120 ± 0.154 & 85.049 ± 0.158 & 85.145 ± 0.157 & 85.542 ± 0.160 & 86.481 ± 0.153 & 86.475 ± 0.156 & 84.471 ± 0.192 \\
4.0 & 85.063 ± 0.157 & 84.993 ± 0.157 & 85.028 ± 0.162 & 85.344 ± 0.154 & 86.447 ± 0.156 & 86.484 ± 0.155 & 84.284 ± 0.202 \\
5.0 & 85.041 ± 0.157 & 84.979 ± 0.163 & 84.986 ± 0.163 & 85.214 ± 0.159 & 86.305 ± 0.156 & 86.462 ± 0.151 & 84.148 ± 0.211 \\
\bottomrule
\end{tabular}
}
\end{table}

\begin{table}[h!]
\centering
\caption{Results for the Arrhythmia dataset.}
{\fontsize{5pt}{7pt}\selectfont
\begin{tabular}{p{0.18cm}|p{1.5153cm}p{1.5153cm}p{1.5153cm}p{1.5153cm}p{1.5153cm}p{1.5153cm}p{1.5153cm}}
\toprule
 BR &     RF(nt\_500) &     RF(qs\_ent) &       RF(mn\_4) &       RF(mn\_8) &       RF(ml\_4) &       RF(ml\_5) &     RF(nf\_all) \\
\midrule
0.2 & 62.300 ± 1.482 & 59.704 ± 1.396 & 62.098 ± 1.578 & 61.287 ± 1.432 & 57.487 ± 0.825 & 56.637 ± 0.439 & 64.712 ± 1.938 \\
0.4 & 68.310 ± 1.629 & 64.137 ± 1.667 & 67.794 ± 1.825 & 66.781 ± 1.821 & 61.971 ± 1.519 & 60.544 ± 1.315 & 73.096 ± 2.006 \\
0.6 & 70.875 ± 1.652 & 66.696 ± 1.631 & 70.719 ± 1.837 & 70.010 ± 1.815 & 65.270 ± 1.722 & 63.580 ± 1.615 & 75.290 ± 1.896 \\
0.8 & 72.371 ± 1.683 & 68.155 ± 1.724 & 72.150 ± 1.771 & 71.561 ± 1.862 & 67.346 ± 1.845 & 65.598 ± 1.804 & 75.999 ± 1.944 \\
1.0 & 73.208 ± 1.767 & 69.061 ± 1.656 & 72.975 ± 1.874 & 72.714 ± 1.928 & 68.610 ± 1.788 & 67.092 ± 1.843 & 76.130 ± 1.929 \\
1.2 & 73.844 ± 1.792 & 69.758 ± 1.631 & 73.538 ± 1.945 & 73.505 ± 1.995 & 70.548 ± 1.793 & 69.507 ± 1.780 & 76.161 ± 1.949 \\
2.0 & 74.887 ± 1.863 & 70.823 ± 1.605 & 74.706 ± 1.863 & 74.654 ± 1.939 & 72.496 ± 1.762 & 71.760 ± 1.868 & 75.277 ± 2.320 \\
3.0 & 75.289 ± 1.910 & 71.452 ± 1.740 & 75.127 ± 1.972 & 75.193 ± 2.043 & 74.018 ± 1.969 & 73.482 ± 1.882 & 74.417 ± 2.534 \\
4.0 & 75.518 ± 1.941 & 71.625 ± 1.742 & 75.327 ± 1.915 & 75.271 ± 1.865 & 74.660 ± 1.920 & 74.217 ± 1.858 & 73.840 ± 2.693 \\
5.0 & 75.554 ± 1.831 & 71.726 ± 1.644 & 75.250 ± 1.948 & 75.427 ± 1.892 & 75.024 ± 1.807 & 74.637 ± 1.956 & 73.299 ± 2.860 \\
\bottomrule
\end{tabular}
}
\end{table}

\begin{table}
\centering
\caption{Results for the Audiology (Standardized) dataset.}
{\fontsize{5pt}{7pt}\selectfont
\begin{tabular}{p{0.18cm}|p{1.5153cm}p{1.5153cm}p{1.5153cm}p{1.5153cm}p{1.5153cm}p{1.5153cm}p{1.5153cm}}
\toprule
 BR &     RF(nt\_500) &     RF(qs\_ent) &       RF(mn\_4) &       RF(mn\_8) &       RF(ml\_4) &       RF(ml\_5) &     RF(nf\_all) \\
\midrule
0.2 & 62.012 ± 3.483 & 60.583 ± 3.708 & 55.974 ± 3.811 & 47.936 ± 2.531 & 41.996 ± 4.582 & 38.258 ± 6.699 & 62.505 ± 3.902 \\
0.4 & 68.927 ± 3.394 & 67.429 ± 3.510 & 66.274 ± 3.570 & 60.223 ± 3.823 & 46.573 ± 2.348 & 45.205 ± 2.154 & 69.445 ± 3.565 \\
0.6 & 71.831 ± 3.390 & 70.062 ± 3.607 & 69.844 ± 3.525 & 65.517 ± 3.301 & 51.308 ± 3.124 & 47.432 ± 2.393 & 72.611 ± 3.894 \\
0.8 & 72.977 ± 3.460 & 71.022 ± 3.488 & 71.411 ± 3.601 & 68.334 ± 3.502 & 56.200 ± 3.591 & 50.707 ± 3.160 & 74.181 ± 4.129 \\
1.0 & 73.600 ± 3.529 & 71.338 ± 3.652 & 72.482 ± 3.604 & 69.635 ± 3.662 & 59.150 ± 3.630 & 53.672 ± 3.550 & 74.912 ± 4.221 \\
1.2 & 73.914 ± 3.642 & 71.940 ± 3.598 & 73.553 ± 3.628 & 71.984 ± 3.495 & 64.130 ± 3.055 & 61.341 ± 3.446 & 75.257 ± 4.040 \\
2.0 & 74.563 ± 3.634 & 72.556 ± 3.819 & 74.459 ± 3.808 & 73.940 ± 3.741 & 67.001 ± 2.997 & 65.144 ± 2.945 & 74.918 ± 4.231 \\
3.0 & 74.742 ± 3.833 & 72.814 ± 3.668 & 74.812 ± 3.897 & 74.913 ± 3.833 & 69.641 ± 3.490 & 68.392 ± 3.249 & 74.473 ± 4.305 \\
4.0 & 75.017 ± 3.864 & 72.908 ± 3.733 & 75.122 ± 3.949 & 75.216 ± 3.679 & 72.248 ± 3.682 & 70.200 ± 3.341 & 74.227 ± 4.441 \\
5.0 & 74.980 ± 3.860 & 72.929 ± 3.729 & 75.037 ± 3.703 & 75.338 ± 3.621 & 73.653 ± 3.646 & 71.930 ± 3.452 & 74.101 ± 4.527 \\
\bottomrule
\end{tabular}
}
\end{table}

\begin{table}[ht]
\centering
\caption{Results for the Australian Credit Approval dataset.}
{\fontsize{5pt}{7pt}\selectfont
\begin{tabular}{p{0.18cm}|p{1.5153cm}p{1.5153cm}p{1.5153cm}p{1.5153cm}p{1.5153cm}p{1.5153cm}p{1.5153cm}}
\toprule
 BR &     RF(nt\_500) &     RF(qs\_ent) &       RF(mn\_4) &       RF(mn\_8) &       RF(ml\_4) &       RF(ml\_5) &     RF(nf\_all) \\
\midrule
0.2 & 86.946 ± 1.397 & 86.680 ± 1.431 & 86.727 ± 1.451 & 86.634 ± 1.430 & 86.312 ± 1.469 & 86.151 ± 1.512 & 86.208 ± 1.450 \\
0.4 & 87.151 ± 1.370 & 86.986 ± 1.429 & 86.974 ± 1.479 & 86.810 ± 1.413 & 86.714 ± 1.419 & 86.564 ± 1.400 & 86.525 ± 1.461 \\
0.6 & 87.225 ± 1.399 & 87.087 ± 1.415 & 87.062 ± 1.422 & 86.945 ± 1.355 & 86.920 ± 1.407 & 86.773 ± 1.429 & 86.541 ± 1.443 \\
0.8 & 87.199 ± 1.411 & 87.098 ± 1.436 & 87.135 ± 1.393 & 87.067 ± 1.400 & 86.947 ± 1.409 & 86.812 ± 1.436 & 86.404 ± 1.490 \\
1.0 & 87.214 ± 1.409 & 87.070 ± 1.430 & 87.089 ± 1.368 & 87.062 ± 1.376 & 86.953 ± 1.409 & 86.901 ± 1.454 & 86.130 ± 1.484 \\
1.2 & 87.188 ± 1.367 & 87.114 ± 1.363 & 87.059 ± 1.393 & 87.055 ± 1.363 & 86.990 ± 1.430 & 86.956 ± 1.371 & 85.850 ± 1.602 \\
2.0 & 87.051 ± 1.356 & 87.018 ± 1.413 & 87.022 ± 1.401 & 87.025 ± 1.412 & 87.105 ± 1.459 & 87.018 ± 1.405 & 84.980 ± 1.749 \\
3.0 & 86.989 ± 1.373 & 86.878 ± 1.381 & 86.917 ± 1.342 & 86.999 ± 1.368 & 87.096 ± 1.366 & 87.104 ± 1.425 & 84.350 ± 1.891 \\
4.0 & 86.944 ± 1.346 & 86.821 ± 1.385 & 86.841 ± 1.417 & 86.888 ± 1.367 & 87.078 ± 1.373 & 87.070 ± 1.410 & 83.967 ± 1.852 \\
5.0 & 86.911 ± 1.375 & 86.824 ± 1.366 & 86.844 ± 1.423 & 86.891 ± 1.388 & 87.073 ± 1.413 & 87.076 ± 1.345 & 83.723 ± 1.931 \\
\bottomrule
\end{tabular}
}
\end{table}

\begin{table}[ht]
\centering
\caption{Results for the Balance Scale dataset.}
{\fontsize{5pt}{7pt}\selectfont
\begin{tabular}{p{0.18cm}|p{1.5153cm}p{1.5153cm}p{1.5153cm}p{1.5153cm}p{1.5153cm}p{1.5153cm}p{1.5153cm}}
\toprule
 BR &     RF(nt\_500) &     RF(qs\_ent) &       RF(mn\_4) &       RF(mn\_8) &       RF(ml\_4) &       RF(ml\_5) &     RF(nf\_all) \\
\midrule
0.2 & 85.972 ± 1.474 & 85.018 ± 1.595 & 85.404 ± 1.599 & 85.519 ± 1.766 & 85.454 ± 1.914 & 85.296 ± 2.031 & 84.374 ± 1.747 \\
0.4 & 84.732 ± 1.452 & 84.095 ± 1.441 & 84.649 ± 1.527 & 84.895 ± 1.719 & 85.386 ± 1.800 & 85.283 ± 1.877 & 82.641 ± 1.830 \\
0.6 & 83.947 ± 1.427 & 83.416 ± 1.440 & 84.227 ± 1.549 & 84.487 ± 1.652 & 84.906 ± 1.668 & 84.959 ± 1.792 & 81.262 ± 1.775 \\
0.8 & 83.329 ± 1.493 & 82.786 ± 1.435 & 83.811 ± 1.464 & 84.268 ± 1.595 & 84.635 ± 1.705 & 84.664 ± 1.765 & 80.093 ± 1.887 \\
1.0 & 82.831 ± 1.449 & 82.333 ± 1.582 & 83.593 ± 1.489 & 84.180 ± 1.585 & 84.490 ± 1.682 & 84.547 ± 1.719 & 79.090 ± 1.913 \\
1.2 & 82.483 ± 1.445 & 81.907 ± 1.519 & 82.909 ± 1.497 & 83.936 ± 1.582 & 84.319 ± 1.667 & 84.341 ± 1.708 & 78.248 ± 1.864 \\
2.0 & 81.581 ± 1.538 & 81.143 ± 1.593 & 81.958 ± 1.603 & 83.419 ± 1.518 & 83.913 ± 1.538 & 84.051 ± 1.590 & 76.734 ± 1.969 \\
3.0 & 81.256 ± 1.579 & 80.743 ± 1.629 & 81.097 ± 1.655 & 82.447 ± 1.562 & 83.619 ± 1.497 & 83.661 ± 1.561 & 76.189 ± 2.062 \\
4.0 & 81.029 ± 1.563 & 80.670 ± 1.628 & 80.938 ± 1.604 & 81.811 ± 1.567 & 83.263 ± 1.442 & 83.512 ± 1.522 & 75.956 ± 2.133 \\
5.0 & 81.015 ± 1.614 & 80.616 ± 1.625 & 80.712 ± 1.610 & 81.326 ± 1.583 & 82.771 ± 1.516 & 83.185 ± 1.564 & 75.788 ± 2.113 \\
\bottomrule
\end{tabular}
}
\end{table}

\begin{table}[ht]
\centering
\caption{Results for the Breast Cancer Wisc. (Diag.) dataset.}
{\fontsize{5pt}{7pt}\selectfont
\begin{tabular}{p{0.18cm}|p{1.5153cm}p{1.5153cm}p{1.5153cm}p{1.5153cm}p{1.5153cm}p{1.5153cm}p{1.5153cm}}
\toprule
 BR &     RF(nt\_500) &     RF(qs\_ent) &       RF(mn\_4) &       RF(mn\_8) &       RF(ml\_4) &       RF(ml\_5) &     RF(nf\_all) \\
\midrule
0.2 & 94.673 ± 1.126 & 94.598 ± 1.201 & 94.460 ± 1.194 & 94.289 ± 1.163 & 93.794 ± 1.273 & 93.567 ± 1.316 & 94.367 ± 1.365 \\
0.4 & 95.076 ± 1.081 & 95.080 ± 1.171 & 94.880 ± 1.112 & 94.664 ± 1.121 & 94.487 ± 1.222 & 94.291 ± 1.277 & 94.704 ± 1.257 \\
0.6 & 95.257 ± 1.111 & 95.314 ± 1.163 & 95.147 ± 1.130 & 94.924 ± 1.117 & 94.730 ± 1.144 & 94.615 ± 1.193 & 94.876 ± 1.313 \\
0.8 & 95.360 ± 1.153 & 95.494 ± 1.144 & 95.241 ± 1.203 & 95.061 ± 1.187 & 94.922 ± 1.164 & 94.746 ± 1.194 & 94.942 ± 1.350 \\
1.0 & 95.459 ± 1.138 & 95.606 ± 1.171 & 95.346 ± 1.140 & 95.122 ± 1.163 & 94.970 ± 1.146 & 94.829 ± 1.206 & 94.930 ± 1.383 \\
1.2 & 95.540 ± 1.165 & 95.675 ± 1.168 & 95.511 ± 1.186 & 95.309 ± 1.152 & 95.176 ± 1.188 & 95.087 ± 1.168 & 94.822 ± 1.371 \\
2.0 & 95.651 ± 1.202 & 95.870 ± 1.110 & 95.710 ± 1.165 & 95.600 ± 1.181 & 95.387 ± 1.183 & 95.265 ± 1.172 & 94.536 ± 1.374 \\
3.0 & 95.724 ± 1.168 & 95.871 ± 1.113 & 95.743 ± 1.188 & 95.642 ± 1.179 & 95.544 ± 1.167 & 95.500 ± 1.163 & 94.256 ± 1.398 \\
4.0 & 95.692 ± 1.172 & 95.897 ± 1.108 & 95.714 ± 1.188 & 95.699 ± 1.186 & 95.607 ± 1.156 & 95.518 ± 1.146 & 94.154 ± 1.462 \\
5.0 & 95.691 ± 1.152 & 95.898 ± 1.095 & 95.702 ± 1.178 & 95.704 ± 1.151 & 95.619 ± 1.166 & 95.550 ± 1.176 & 94.007 ± 1.442 \\
\bottomrule
\end{tabular}
}
\end{table}

\begin{table}[ht]
\centering
\caption{Results for the Breast Cancer Wisc. (Orig.) dataset.}
{\fontsize{5pt}{7pt}\selectfont
\begin{tabular}{p{0.18cm}|p{1.5153cm}p{1.5153cm}p{1.5153cm}p{1.5153cm}p{1.5153cm}p{1.5153cm}p{1.5153cm}}
\toprule
 BR &     RF(nt\_500) &     RF(qs\_ent) &       RF(mn\_4) &       RF(mn\_8) &       RF(ml\_4) &       RF(ml\_5) &     RF(nf\_all) \\
\midrule
0.2 & 95.366 ± 1.105 & 95.212 ± 1.144 & 95.321 ± 1.128 & 95.454 ± 1.123 & 95.458 ± 1.128 & 95.423 ± 1.112 & 94.517 ± 1.298 \\
0.4 & 95.506 ± 1.085 & 95.309 ± 1.113 & 95.350 ± 1.113 & 95.474 ± 1.142 & 95.422 ± 1.160 & 95.386 ± 1.200 & 94.552 ± 1.255 \\
0.6 & 95.491 ± 1.085 & 95.362 ± 1.127 & 95.440 ± 1.112 & 95.501 ± 1.118 & 95.435 ± 1.138 & 95.415 ± 1.170 & 94.570 ± 1.286 \\
0.8 & 95.457 ± 1.070 & 95.342 ± 1.135 & 95.403 ± 1.103 & 95.440 ± 1.122 & 95.411 ± 1.124 & 95.376 ± 1.133 & 94.521 ± 1.274 \\
1.0 & 95.429 ± 1.055 & 95.287 ± 1.087 & 95.355 ± 1.097 & 95.433 ± 1.116 & 95.401 ± 1.099 & 95.382 ± 1.105 & 94.407 ± 1.309 \\
1.2 & 95.378 ± 1.077 & 95.247 ± 1.081 & 95.312 ± 1.109 & 95.394 ± 1.120 & 95.393 ± 1.130 & 95.403 ± 1.118 & 94.336 ± 1.357 \\
2.0 & 95.262 ± 1.086 & 95.162 ± 1.109 & 95.176 ± 1.112 & 95.233 ± 1.095 & 95.341 ± 1.111 & 95.324 ± 1.108 & 93.919 ± 1.473 \\
3.0 & 95.166 ± 1.105 & 95.133 ± 1.110 & 95.091 ± 1.124 & 95.151 ± 1.142 & 95.249 ± 1.138 & 95.279 ± 1.174 & 93.516 ± 1.569 \\
4.0 & 95.155 ± 1.134 & 95.102 ± 1.124 & 95.026 ± 1.140 & 95.089 ± 1.124 & 95.179 ± 1.132 & 95.207 ± 1.145 & 93.316 ± 1.590 \\
5.0 & 95.115 ± 1.120 & 95.064 ± 1.128 & 95.024 ± 1.153 & 95.047 ± 1.133 & 95.145 ± 1.104 & 95.183 ± 1.107 & 93.100 ± 1.618 \\
\bottomrule
\end{tabular}
}
\end{table}

\begin{table}[ht]
\centering
\caption{Results for the Congressional Voting Rec. dataset.}
{\fontsize{5pt}{7pt}\selectfont
\begin{tabular}{p{0.18cm}|p{1.5153cm}p{1.5153cm}p{1.5153cm}p{1.5153cm}p{1.5153cm}p{1.5153cm}p{1.5153cm}}
\toprule
 BR &     RF(nt\_500) &     RF(qs\_ent) &       RF(mn\_4) &       RF(mn\_8) &       RF(ml\_4) &       RF(ml\_5) &     RF(nf\_all) \\
\midrule
0.2 & 93.690 ± 1.487 & 93.918 ± 1.474 & 93.819 ± 1.437 & 93.687 ± 1.439 & 92.997 ± 1.550 & 92.795 ± 1.658 & 94.404 ± 1.337 \\
0.4 & 94.355 ± 1.520 & 94.412 ± 1.509 & 94.278 ± 1.471 & 93.974 ± 1.410 & 93.683 ± 1.421 & 93.534 ± 1.469 & 94.373 ± 1.394 \\
0.6 & 94.592 ± 1.480 & 94.640 ± 1.489 & 94.509 ± 1.517 & 94.202 ± 1.429 & 93.807 ± 1.380 & 93.747 ± 1.428 & 94.301 ± 1.448 \\
0.8 & 94.661 ± 1.494 & 94.692 ± 1.471 & 94.637 ± 1.519 & 94.377 ± 1.483 & 93.886 ± 1.462 & 93.800 ± 1.449 & 94.156 ± 1.453 \\
1.0 & 94.702 ± 1.431 & 94.683 ± 1.451 & 94.686 ± 1.490 & 94.512 ± 1.500 & 94.028 ± 1.489 & 93.890 ± 1.455 & 93.915 ± 1.579 \\
1.2 & 94.681 ± 1.457 & 94.662 ± 1.442 & 94.763 ± 1.480 & 94.652 ± 1.504 & 94.276 ± 1.524 & 94.099 ± 1.484 & 93.702 ± 1.684 \\
2.0 & 94.675 ± 1.426 & 94.604 ± 1.443 & 94.753 ± 1.455 & 94.795 ± 1.474 & 94.481 ± 1.540 & 94.317 ± 1.505 & 93.139 ± 1.782 \\
3.0 & 94.620 ± 1.413 & 94.569 ± 1.389 & 94.605 ± 1.457 & 94.760 ± 1.461 & 94.705 ± 1.485 & 94.620 ± 1.515 & 92.788 ± 1.878 \\
4.0 & 94.615 ± 1.413 & 94.539 ± 1.423 & 94.575 ± 1.394 & 94.703 ± 1.433 & 94.756 ± 1.501 & 94.722 ± 1.513 & 92.721 ± 1.931 \\
5.0 & 94.621 ± 1.432 & 94.520 ± 1.451 & 94.520 ± 1.404 & 94.633 ± 1.425 & 94.791 ± 1.472 & 94.769 ± 1.483 & 92.645 ± 1.922 \\
\bottomrule
\end{tabular}
}
\end{table}

\begin{table}[ht]
\centering
\caption{Results for the Echocardiogram dataset.}
{\fontsize{5pt}{7pt}\selectfont
\begin{tabular}{p{0.18cm}|p{1.5153cm}p{1.5153cm}p{1.5153cm}p{1.5153cm}p{1.5153cm}p{1.5153cm}p{1.5153cm}}
\toprule
 BR &     RF(nt\_500) &     RF(qs\_ent) &       RF(mn\_4) &       RF(mn\_8) &       RF(ml\_4) &       RF(ml\_5) &     RF(nf\_all) \\
\midrule
0.2 & 70.774 ± 2.633 & 70.992 ± 2.462 & 71.153 ± 2.174 & 70.968 ± 0.000 & 70.968 ± 0.000 & 70.968 ± 0.000 & 71.226 ± 3.311 \\
0.4 & 71.242 ± 4.116 & 71.435 ± 3.788 & 71.524 ± 3.908 & 71.427 ± 3.500 & 70.944 ± 0.665 & 70.968 ± 0.000 & 71.718 ± 4.577 \\
0.6 & 72.024 ± 4.493 & 72.468 ± 4.497 & 72.500 ± 4.746 & 72.315 ± 4.755 & 71.766 ± 3.347 & 71.056 ± 2.374 & 72.000 ± 5.264 \\
0.8 & 71.790 ± 5.039 & 72.258 ± 5.018 & 72.524 ± 4.697 & 72.581 ± 4.823 & 72.444 ± 3.855 & 71.944 ± 3.446 & 70.774 ± 6.117 \\
1.0 & 71.831 ± 4.925 & 72.298 ± 5.077 & 72.306 ± 5.037 & 72.685 ± 4.923 & 72.645 ± 4.224 & 72.403 ± 3.853 & 70.500 ± 6.162 \\
1.2 & 71.766 ± 5.117 & 72.008 ± 5.478 & 72.008 ± 5.353 & 72.718 ± 5.119 & 73.048 ± 4.601 & 72.976 ± 4.423 & 69.919 ± 6.735 \\
2.0 & 71.589 ± 5.629 & 71.581 ± 6.115 & 71.565 ± 5.413 & 72.331 ± 5.516 & 72.887 ± 5.019 & 73.113 ± 5.005 & 68.355 ± 7.291 \\
3.0 & 71.387 ± 5.876 & 71.927 ± 6.038 & 71.266 ± 5.837 & 72.194 ± 5.717 & 72.935 ± 5.458 & 73.065 ± 5.224 & 68.105 ± 7.742 \\
4.0 & 71.419 ± 6.142 & 71.460 ± 5.895 & 71.177 ± 5.918 & 71.403 ± 5.795 & 72.540 ± 5.775 & 72.815 ± 5.564 & 67.847 ± 7.604 \\
5.0 & 71.234 ± 6.367 & 71.734 ± 6.113 & 71.161 ± 6.173 & 71.387 ± 5.916 & 72.234 ± 5.867 & 72.452 ± 5.595 & 67.419 ± 8.158 \\
\bottomrule
\end{tabular}
}
\end{table}

\begin{table}[ht]
\centering
\caption{Results for the Ecoli dataset.}
{\fontsize{5pt}{7pt}\selectfont
\begin{tabular}{p{0.18cm}|p{1.5153cm}p{1.5153cm}p{1.5153cm}p{1.5153cm}p{1.5153cm}p{1.5153cm}p{1.5153cm}}
\toprule
 BR &     RF(nt\_500) &     RF(qs\_ent) &       RF(mn\_4) &       RF(mn\_8) &       RF(ml\_4) &       RF(ml\_5) &     RF(nf\_all) \\
\midrule
0.2 & 84.060 ± 1.965 & 83.714 ± 2.027 & 83.295 ± 2.017 & 79.817 ± 2.070 & 76.765 ± 1.462 & 76.155 ± 1.311 & 84.161 ± 1.945 \\
0.4 & 85.527 ± 1.799 & 84.943 ± 1.886 & 85.220 ± 1.942 & 84.589 ± 1.864 & 81.098 ± 1.966 & 78.716 ± 1.762 & 84.644 ± 1.994 \\
0.6 & 85.835 ± 1.858 & 85.263 ± 1.928 & 85.680 ± 1.843 & 85.378 ± 1.922 & 83.496 ± 1.856 & 82.314 ± 1.962 & 84.339 ± 2.127 \\
0.8 & 85.671 ± 1.872 & 85.164 ± 1.926 & 85.580 ± 2.000 & 85.452 ± 1.914 & 84.201 ± 1.834 & 83.573 ± 1.996 & 83.664 ± 2.242 \\
1.0 & 85.565 ± 1.905 & 85.082 ± 1.949 & 85.589 ± 1.975 & 85.509 ± 1.924 & 84.560 ± 1.868 & 84.170 ± 1.912 & 83.049 ± 2.438 \\
1.2 & 85.321 ± 1.981 & 84.933 ± 1.996 & 85.592 ± 1.928 & 85.554 ± 1.939 & 85.046 ± 1.895 & 84.647 ± 1.905 & 82.658 ± 2.535 \\
2.0 & 84.859 ± 2.041 & 84.394 ± 2.073 & 85.006 ± 2.060 & 85.351 ± 2.069 & 85.106 ± 1.968 & 84.902 ± 1.945 & 81.269 ± 2.666 \\
3.0 & 84.604 ± 2.054 & 84.104 ± 2.132 & 84.621 ± 2.049 & 84.946 ± 2.020 & 85.171 ± 2.015 & 85.070 ± 1.976 & 80.528 ± 2.765 \\
4.0 & 84.490 ± 2.110 & 83.978 ± 2.036 & 84.439 ± 2.111 & 84.690 ± 2.096 & 85.019 ± 2.050 & 85.054 ± 1.999 & 80.137 ± 2.788 \\
5.0 & 84.369 ± 2.111 & 83.912 ± 2.060 & 84.348 ± 2.103 & 84.568 ± 2.130 & 84.876 ± 2.011 & 84.990 ± 2.051 & 79.839 ± 2.765 \\
\bottomrule
\end{tabular}
}
\end{table}

\begin{table}[ht]
\centering
\caption{Results for the German Credit Data dataset.}
{\fontsize{5pt}{7pt}\selectfont
\begin{tabular}{p{0.18cm}|p{1.5153cm}p{1.5153cm}p{1.5153cm}p{1.5153cm}p{1.5153cm}p{1.5153cm}p{1.5153cm}}
\toprule
 BR &     RF(nt\_500) &     RF(qs\_ent) &       RF(mn\_4) &       RF(mn\_8) &       RF(ml\_4) &       RF(ml\_5) &     RF(nf\_all) \\
\midrule
0.2 & 73.932 ± 1.059 & 73.797 ± 1.169 & 73.614 ± 1.169 & 73.195 ± 1.066 & 71.769 ± 0.839 & 71.227 ± 0.738 & 74.612 ± 1.322 \\
0.4 & 75.023 ± 1.137 & 74.778 ± 1.190 & 74.700 ± 1.159 & 74.337 ± 1.176 & 73.230 ± 1.031 & 72.766 ± 0.938 & 75.046 ± 1.416 \\
0.6 & 75.312 ± 1.110 & 75.072 ± 1.172 & 74.996 ± 1.262 & 74.816 ± 1.201 & 73.879 ± 1.090 & 73.483 ± 1.084 & 74.919 ± 1.423 \\
0.8 & 75.410 ± 1.149 & 75.157 ± 1.199 & 75.127 ± 1.222 & 74.954 ± 1.223 & 74.276 ± 1.137 & 73.908 ± 1.113 & 74.567 ± 1.524 \\
1.0 & 75.397 ± 1.150 & 75.195 ± 1.232 & 75.170 ± 1.197 & 75.174 ± 1.190 & 74.510 ± 1.164 & 74.178 ± 1.175 & 74.344 ± 1.615 \\
1.2 & 75.466 ± 1.196 & 75.296 ± 1.272 & 75.262 ± 1.240 & 75.153 ± 1.242 & 74.798 ± 1.168 & 74.636 ± 1.164 & 74.127 ± 1.597 \\
2.0 & 75.352 ± 1.196 & 75.113 ± 1.263 & 75.164 ± 1.280 & 75.154 ± 1.209 & 75.114 ± 1.217 & 75.035 ± 1.226 & 73.342 ± 1.655 \\
3.0 & 75.375 ± 1.211 & 75.078 ± 1.209 & 75.082 ± 1.299 & 75.164 ± 1.303 & 75.324 ± 1.197 & 75.207 ± 1.191 & 72.835 ± 1.749 \\
4.0 & 75.349 ± 1.226 & 75.073 ± 1.302 & 75.007 ± 1.288 & 75.124 ± 1.263 & 75.252 ± 1.256 & 75.249 ± 1.225 & 72.440 ± 1.781 \\
5.0 & 75.322 ± 1.218 & 75.106 ± 1.280 & 75.110 ± 1.244 & 75.178 ± 1.279 & 75.216 ± 1.278 & 75.245 ± 1.348 & 72.175 ± 1.742 \\
\bottomrule
\end{tabular}
}
\end{table}

\begin{table}[ht]
\centering
\caption{Results for the Glass Identification dataset.}
{\fontsize{5pt}{7pt}\selectfont
\begin{tabular}{p{0.18cm}|p{1.5153cm}p{1.5153cm}p{1.5153cm}p{1.5153cm}p{1.5153cm}p{1.5153cm}p{1.5153cm}}
\toprule
 BR &     RF(nt\_500) &     RF(qs\_ent) &       RF(mn\_4) &       RF(mn\_8) &       RF(ml\_4) &       RF(ml\_5) &     RF(nf\_all) \\
\midrule
0.2 & 67.159 ± 3.839 & 67.054 ± 3.664 & 65.267 ± 3.704 & 63.493 ± 3.843 & 61.211 ± 3.696 & 60.005 ± 3.950 & 65.789 ± 3.919 \\
0.4 & 72.204 ± 3.639 & 72.366 ± 3.629 & 70.746 ± 3.679 & 67.456 ± 3.605 & 65.249 ± 3.781 & 64.087 ± 3.939 & 71.043 ± 3.915 \\
0.6 & 74.004 ± 3.690 & 74.222 ± 3.719 & 72.821 ± 3.492 & 70.084 ± 3.738 & 67.529 ± 3.871 & 66.000 ± 3.737 & 72.572 ± 3.927 \\
0.8 & 74.934 ± 3.675 & 74.812 ± 3.579 & 73.941 ± 3.784 & 71.793 ± 3.779 & 69.305 ± 3.829 & 67.435 ± 3.690 & 72.891 ± 3.877 \\
1.0 & 75.250 ± 3.652 & 75.154 ± 3.608 & 74.346 ± 3.865 & 72.728 ± 3.859 & 70.676 ± 3.851 & 68.794 ± 3.761 & 72.829 ± 3.958 \\
1.2 & 75.413 ± 3.695 & 75.424 ± 3.659 & 74.727 ± 3.768 & 73.748 ± 3.932 & 72.845 ± 3.870 & 71.479 ± 3.804 & 72.798 ± 3.983 \\
2.0 & 75.577 ± 3.698 & 75.596 ± 3.778 & 75.176 ± 3.813 & 74.460 ± 3.889 & 74.366 ± 3.795 & 73.659 ± 3.823 & 71.466 ± 4.357 \\
3.0 & 75.429 ± 3.751 & 75.383 ± 3.969 & 74.927 ± 3.667 & 74.819 ± 3.593 & 74.840 ± 3.687 & 74.624 ± 3.789 & 70.340 ± 4.430 \\
4.0 & 75.328 ± 3.788 & 75.457 ± 3.842 & 74.848 ± 3.641 & 74.708 ± 3.744 & 74.868 ± 3.641 & 74.725 ± 3.698 & 69.839 ± 4.622 \\
5.0 & 75.295 ± 3.721 & 75.363 ± 3.920 & 74.923 ± 3.736 & 74.838 ± 3.877 & 75.037 ± 3.738 & 74.931 ± 3.725 & 69.432 ± 4.705 \\
\bottomrule
\end{tabular}
}
\end{table}

\begin{table}[ht]
\centering
\caption{Results for the Heart dataset.}
{\fontsize{5pt}{7pt}\selectfont
\begin{tabular}{p{0.18cm}|p{1.5153cm}p{1.5153cm}p{1.5153cm}p{1.5153cm}p{1.5153cm}p{1.5153cm}p{1.5153cm}}
\toprule
 BR &     RF(nt\_500) &     RF(qs\_ent) &       RF(mn\_4) &       RF(mn\_8) &       RF(ml\_4) &       RF(ml\_5) &     RF(nf\_all) \\
\midrule
0.2 & 83.319 ± 2.532 & 82.794 ± 2.558 & 82.969 ± 2.641 & 82.924 ± 2.685 & 83.324 ± 2.569 & 83.217 ± 2.502 & 81.669 ± 2.773 \\
0.4 & 82.693 ± 2.339 & 82.406 ± 2.480 & 82.696 ± 2.461 & 82.806 ± 2.542 & 83.294 ± 2.553 & 83.224 ± 2.662 & 81.289 ± 2.810 \\
0.6 & 82.272 ± 2.432 & 82.019 ± 2.495 & 82.243 ± 2.514 & 82.472 ± 2.505 & 83.137 ± 2.650 & 83.119 ± 2.584 & 80.676 ± 2.961 \\
0.8 & 81.907 ± 2.465 & 81.800 ± 2.614 & 82.061 ± 2.490 & 82.352 ± 2.558 & 82.981 ± 2.534 & 83.148 ± 2.617 & 80.409 ± 2.981 \\
1.0 & 81.689 ± 2.472 & 81.639 ± 2.565 & 81.846 ± 2.545 & 82.272 ± 2.470 & 82.933 ± 2.555 & 82.993 ± 2.525 & 79.961 ± 3.096 \\
1.2 & 81.506 ± 2.501 & 81.372 ± 2.551 & 81.578 ± 2.623 & 81.917 ± 2.569 & 82.594 ± 2.550 & 82.830 ± 2.556 & 79.502 ± 2.959 \\
2.0 & 81.046 ± 2.466 & 80.894 ± 2.481 & 80.946 ± 2.510 & 81.361 ± 2.428 & 82.093 ± 2.390 & 82.435 ± 2.453 & 78.394 ± 3.078 \\
3.0 & 80.854 ± 2.523 & 80.707 ± 2.517 & 80.744 ± 2.465 & 81.083 ± 2.619 & 81.750 ± 2.612 & 81.989 ± 2.632 & 77.691 ± 3.164 \\
4.0 & 80.813 ± 2.517 & 80.639 ± 2.563 & 80.680 ± 2.578 & 80.926 ± 2.610 & 81.441 ± 2.609 & 81.539 ± 2.475 & 77.250 ± 3.278 \\
5.0 & 80.774 ± 2.541 & 80.691 ± 2.510 & 80.583 ± 2.552 & 80.835 ± 2.582 & 81.180 ± 2.652 & 81.322 ± 2.607 & 77.019 ± 3.233 \\
\bottomrule
\end{tabular}
}
\end{table}

\begin{table}[ht]
\centering
\caption{Results for the Hepatitis dataset.}
{\fontsize{5pt}{7pt}\selectfont
\begin{tabular}{p{0.18cm}|p{1.5153cm}p{1.5153cm}p{1.5153cm}p{1.5153cm}p{1.5153cm}p{1.5153cm}p{1.5153cm}}
\toprule
 BR &     RF(nt\_500) &     RF(qs\_ent) &       RF(mn\_4) &       RF(mn\_8) &       RF(ml\_4) &       RF(ml\_5) &     RF(nf\_all) \\
\midrule
0.2 & 83.399 ± 2.143 & 82.936 ± 2.195 & 82.858 ± 2.107 & 82.081 ± 1.879 & 79.730 ± 0.000 & 79.730 ± 0.000 & 84.115 ± 2.303 \\
0.4 & 84.726 ± 2.600 & 84.287 ± 2.784 & 84.149 ± 2.513 & 83.760 ± 2.454 & 80.831 ± 1.246 & 79.943 ± 0.501 & 84.044 ± 3.243 \\
0.6 & 84.561 ± 2.587 & 84.446 ± 2.868 & 84.220 ± 2.789 & 84.189 ± 2.575 & 82.530 ± 1.869 & 81.128 ± 1.384 & 83.645 ± 3.406 \\
0.8 & 84.355 ± 2.823 & 84.243 ± 2.997 & 84.270 ± 2.976 & 84.226 ± 2.794 & 83.439 ± 2.194 & 82.537 ± 1.866 & 83.233 ± 3.526 \\
1.0 & 84.318 ± 2.889 & 84.216 ± 3.067 & 83.993 ± 3.007 & 84.176 ± 2.886 & 83.838 ± 2.281 & 83.236 ± 2.111 & 82.720 ± 3.387 \\
1.2 & 84.341 ± 2.899 & 84.274 ± 2.999 & 83.993 ± 3.004 & 84.108 ± 2.986 & 84.206 ± 2.530 & 83.905 ± 2.335 & 82.274 ± 3.660 \\
2.0 & 84.111 ± 3.048 & 84.095 ± 3.154 & 84.020 ± 3.136 & 83.970 ± 3.050 & 84.166 ± 2.772 & 84.169 ± 2.627 & 81.081 ± 3.963 \\
3.0 & 84.020 ± 3.081 & 84.061 ± 3.223 & 83.760 ± 3.168 & 83.905 ± 3.138 & 84.020 ± 2.928 & 84.068 ± 2.862 & 80.436 ± 4.001 \\
4.0 & 83.976 ± 3.042 & 83.895 ± 3.169 & 83.723 ± 3.068 & 83.851 ± 3.163 & 84.027 ± 2.988 & 83.983 ± 2.959 & 80.068 ± 4.158 \\
5.0 & 83.983 ± 3.152 & 83.922 ± 3.131 & 83.672 ± 3.239 & 83.696 ± 3.054 & 83.814 ± 3.100 & 83.929 ± 2.956 & 79.959 ± 4.031 \\
\bottomrule
\end{tabular}
}
\end{table}

\begin{table}[ht]
\centering
\caption{Results for the Horse Colic dataset.}
{\fontsize{5pt}{7pt}\selectfont
\begin{tabular}{p{0.18cm}|p{1.5153cm}p{1.5153cm}p{1.5153cm}p{1.5153cm}p{1.5153cm}p{1.5153cm}p{1.5153cm}}
\toprule
 BR &     RF(nt\_500) &     RF(qs\_ent) &       RF(mn\_4) &       RF(mn\_8) &       RF(ml\_4) &       RF(ml\_5) &     RF(nf\_all) \\
\midrule
0.2 & 85.883 ± 1.902 & 85.681 ± 1.959 & 85.645 ± 1.848 & 85.685 ± 1.904 & 84.041 ± 2.054 & 82.474 ± 2.331 & 84.861 ± 2.541 \\
0.4 & 86.111 ± 1.882 & 85.933 ± 1.905 & 86.056 ± 1.812 & 86.090 ± 1.837 & 85.670 ± 1.840 & 85.458 ± 1.822 & 86.052 ± 2.245 \\
0.6 & 86.285 ± 1.872 & 86.240 ± 1.858 & 86.193 ± 1.855 & 86.230 ± 1.843 & 85.780 ± 1.860 & 85.698 ± 1.878 & 86.269 ± 2.141 \\
0.8 & 86.484 ± 1.879 & 86.279 ± 1.903 & 86.315 ± 1.932 & 86.216 ± 1.881 & 85.815 ± 1.907 & 85.774 ± 1.798 & 86.095 ± 2.068 \\
1.0 & 86.516 ± 1.918 & 86.311 ± 1.852 & 86.303 ± 1.808 & 86.197 ± 1.805 & 85.818 ± 1.927 & 85.766 ± 1.826 & 85.769 ± 2.199 \\
1.2 & 86.515 ± 1.888 & 86.365 ± 1.855 & 86.371 ± 1.910 & 86.295 ± 1.827 & 85.856 ± 1.839 & 85.780 ± 1.875 & 85.387 ± 2.231 \\
2.0 & 86.438 ± 1.871 & 86.226 ± 1.920 & 86.360 ± 1.873 & 86.383 ± 1.901 & 85.887 ± 1.998 & 85.781 ± 1.962 & 84.645 ± 2.301 \\
3.0 & 86.480 ± 1.872 & 86.202 ± 1.919 & 86.306 ± 1.886 & 86.421 ± 1.902 & 86.144 ± 1.953 & 85.950 ± 1.870 & 84.007 ± 2.395 \\
4.0 & 86.395 ± 1.844 & 86.148 ± 1.928 & 86.156 ± 1.915 & 86.205 ± 1.865 & 86.167 ± 1.926 & 86.163 ± 1.899 & 83.776 ± 2.420 \\
5.0 & 86.414 ± 1.846 & 86.177 ± 1.797 & 86.261 ± 1.926 & 86.276 ± 1.861 & 86.216 ± 1.936 & 86.166 ± 1.941 & 83.569 ± 2.373 \\
\bottomrule
\end{tabular}
}
\end{table}

\begin{table}[ht]
\centering
\caption{Results for the Image Segmentation (Stat.) dataset.}
{\fontsize{5pt}{7pt}\selectfont
\begin{tabular}{p{0.18cm}|p{1.5153cm}p{1.5153cm}p{1.5153cm}p{1.5153cm}p{1.5153cm}p{1.5153cm}p{1.5153cm}}
\toprule
 BR &     RF(nt\_500) &     RF(qs\_ent) &       RF(mn\_4) &       RF(mn\_8) &       RF(ml\_4) &       RF(ml\_5) &     RF(nf\_all) \\
\midrule
0.2 & 95.150 ± 0.754 & 95.081 ± 0.718 & 94.914 ± 0.783 & 94.360 ± 0.816 & 93.540 ± 0.785 & 93.179 ± 0.774 & 95.126 ± 0.739 \\
0.4 & 96.213 ± 0.583 & 96.143 ± 0.605 & 96.049 ± 0.615 & 95.634 ± 0.679 & 94.868 ± 0.759 & 94.518 ± 0.798 & 95.859 ± 0.662 \\
0.6 & 96.511 ± 0.539 & 96.502 ± 0.577 & 96.418 ± 0.555 & 96.130 ± 0.581 & 95.465 ± 0.681 & 95.151 ± 0.722 & 96.070 ± 0.612 \\
0.8 & 96.712 ± 0.528 & 96.718 ± 0.559 & 96.608 ± 0.548 & 96.365 ± 0.547 & 95.802 ± 0.645 & 95.519 ± 0.665 & 96.150 ± 0.621 \\
1.0 & 96.825 ± 0.509 & 96.843 ± 0.563 & 96.728 ± 0.544 & 96.498 ± 0.564 & 95.996 ± 0.625 & 95.720 ± 0.663 & 96.184 ± 0.631 \\
1.2 & 96.898 ± 0.513 & 96.924 ± 0.522 & 96.814 ± 0.528 & 96.680 ± 0.549 & 96.328 ± 0.589 & 96.104 ± 0.618 & 96.190 ± 0.625 \\
2.0 & 97.018 ± 0.512 & 97.071 ± 0.539 & 96.967 ± 0.524 & 96.876 ± 0.529 & 96.604 ± 0.552 & 96.479 ± 0.548 & 96.094 ± 0.662 \\
3.0 & 97.064 ± 0.525 & 97.107 ± 0.522 & 96.989 ± 0.532 & 96.993 ± 0.524 & 96.805 ± 0.548 & 96.702 ± 0.553 & 95.970 ± 0.669 \\
4.0 & 97.081 ± 0.521 & 97.130 ± 0.544 & 97.023 ± 0.523 & 97.017 ± 0.537 & 96.906 ± 0.539 & 96.824 ± 0.525 & 95.842 ± 0.690 \\
5.0 & 97.080 ± 0.530 & 97.133 ± 0.538 & 97.018 ± 0.538 & 97.013 ± 0.544 & 96.939 ± 0.540 & 96.880 ± 0.554 & 95.773 ± 0.692 \\
\bottomrule
\end{tabular}
}
\end{table}

\begin{table}[ht]
\centering
\caption{Results for the Ionosphere dataset.}
{\fontsize{5pt}{7pt}\selectfont
\begin{tabular}{p{0.18cm}|p{1.5153cm}p{1.5153cm}p{1.5153cm}p{1.5153cm}p{1.5153cm}p{1.5153cm}p{1.5153cm}}
\toprule
 BR &     RF(nt\_500) &     RF(qs\_ent) &       RF(mn\_4) &       RF(mn\_8) &       RF(ml\_4) &       RF(ml\_5) &     RF(nf\_all) \\
\midrule
0.2 & 92.629 ± 1.511 & 91.544 ± 1.839 & 91.913 ± 1.651 & 91.934 ± 1.701 & 88.669 ± 2.406 & 86.764 ± 2.230 & 91.700 ± 1.979 \\
0.4 & 93.076 ± 1.448 & 92.580 ± 1.591 & 92.643 ± 1.533 & 92.511 ± 1.575 & 91.760 ± 1.758 & 91.183 ± 1.996 & 92.256 ± 2.013 \\
0.6 & 93.194 ± 1.495 & 92.753 ± 1.513 & 92.699 ± 1.550 & 92.666 ± 1.538 & 92.253 ± 1.611 & 91.986 ± 1.685 & 92.190 ± 2.083 \\
0.8 & 93.209 ± 1.475 & 92.937 ± 1.544 & 92.829 ± 1.547 & 92.707 ± 1.561 & 92.443 ± 1.574 & 92.310 ± 1.635 & 91.961 ± 2.139 \\
1.0 & 93.231 ± 1.524 & 93.024 ± 1.499 & 92.881 ± 1.493 & 92.764 ± 1.564 & 92.591 ± 1.587 & 92.400 ± 1.593 & 91.730 ± 2.169 \\
1.2 & 93.254 ± 1.526 & 93.020 ± 1.588 & 92.903 ± 1.637 & 92.814 ± 1.613 & 92.776 ± 1.586 & 92.630 ± 1.593 & 91.571 ± 2.190 \\
2.0 & 93.233 ± 1.595 & 93.070 ± 1.556 & 92.957 ± 1.640 & 92.851 ± 1.627 & 92.843 ± 1.572 & 92.770 ± 1.592 & 90.896 ± 2.281 \\
3.0 & 93.229 ± 1.600 & 93.099 ± 1.608 & 93.003 ± 1.614 & 92.906 ± 1.615 & 92.897 ± 1.594 & 92.836 ± 1.572 & 90.474 ± 2.443 \\
4.0 & 93.240 ± 1.598 & 93.103 ± 1.523 & 93.010 ± 1.632 & 92.956 ± 1.709 & 92.844 ± 1.627 & 92.881 ± 1.597 & 90.151 ± 2.442 \\
5.0 & 93.237 ± 1.571 & 93.127 ± 1.577 & 93.039 ± 1.615 & 92.964 ± 1.625 & 92.911 ± 1.604 & 92.927 ± 1.559 & 89.920 ± 2.498 \\
\bottomrule
\end{tabular}
}
\end{table}

\begin{table}[ht]
\centering
\caption{Results for the Iris dataset.}
{\fontsize{5pt}{7pt}\selectfont
\begin{tabular}{p{0.18cm}|p{1.5153cm}p{1.5153cm}p{1.5153cm}p{1.5153cm}p{1.5153cm}p{1.5153cm}p{1.5153cm}}
\toprule
 BR &     RF(nt\_500) &     RF(qs\_ent) &       RF(mn\_4) &       RF(mn\_8) &       RF(ml\_4) &       RF(ml\_5) &     RF(nf\_all) \\
\midrule
0.2 & 95.178 ± 1.824 & 95.098 ± 1.833 & 95.171 ± 1.876 & 95.199 ± 1.899 & 95.095 ± 1.951 & 94.880 ± 2.076 & 95.218 ± 1.840 \\
0.4 & 95.115 ± 1.831 & 95.058 ± 1.804 & 95.232 ± 1.900 & 95.185 ± 1.878 & 95.155 ± 1.835 & 95.212 ± 1.860 & 95.084 ± 1.856 \\
0.6 & 95.010 ± 1.870 & 94.947 ± 1.863 & 95.145 ± 1.941 & 95.185 ± 1.983 & 95.081 ± 1.923 & 95.081 ± 1.949 & 95.017 ± 1.901 \\
0.8 & 94.967 ± 1.892 & 94.859 ± 1.883 & 95.091 ± 1.982 & 95.131 ± 2.049 & 94.963 ± 1.986 & 94.947 ± 1.980 & 94.940 ± 1.926 \\
1.0 & 94.950 ± 1.904 & 94.863 ± 1.915 & 95.064 ± 1.972 & 95.101 ± 2.033 & 94.920 ± 2.030 & 94.859 ± 1.989 & 94.886 ± 1.947 \\
1.2 & 94.917 ± 1.890 & 94.860 ± 1.931 & 95.020 ± 1.977 & 95.138 ± 2.050 & 94.919 ± 2.036 & 94.896 ± 2.015 & 94.883 ± 1.989 \\
2.0 & 94.863 ± 1.919 & 94.826 ± 1.915 & 94.933 ± 1.942 & 95.155 ± 2.053 & 94.943 ± 2.067 & 94.876 ± 2.039 & 94.624 ± 2.239 \\
3.0 & 94.883 ± 1.893 & 94.819 ± 1.865 & 94.842 ± 1.929 & 95.054 ± 1.989 & 94.947 ± 2.000 & 95.001 ± 2.102 & 94.517 ± 2.294 \\
4.0 & 94.839 ± 1.898 & 94.816 ± 1.928 & 94.812 ± 1.916 & 94.964 ± 1.959 & 94.897 ± 1.976 & 94.954 ± 2.050 & 94.517 ± 2.223 \\
5.0 & 94.849 ± 1.914 & 94.806 ± 1.895 & 94.792 ± 1.899 & 94.890 ± 1.955 & 94.833 ± 1.946 & 94.897 ± 1.981 & 94.500 ± 2.237 \\
\bottomrule
\end{tabular}
}
\end{table}

\begin{table}[ht]
\centering
\caption{Results for the Labor Relations dataset.}
{\fontsize{5pt}{7pt}\selectfont
\begin{tabular}{p{0.18cm}|p{1.5153cm}p{1.5153cm}p{1.5153cm}p{1.5153cm}p{1.5153cm}p{1.5153cm}p{1.5153cm}}
\toprule
 BR &     RF(nt\_500) &     RF(qs\_ent) &       RF(mn\_4) &       RF(mn\_8) &       RF(ml\_4) &       RF(ml\_5) &     RF(nf\_all) \\
\midrule
0.2 & 82.500 ± 5.787 & 83.768 ± 6.384 & 80.735 ± 6.456 & 64.901 ± 0.616 & 64.901 ± 0.616 & 64.901 ± 0.616 & 84.817 ± 6.153 \\
0.4 & 90.273 ± 5.306 & 90.113 ± 5.135 & 89.246 ± 5.120 & 84.780 ± 5.871 & 66.436 ± 2.657 & 64.901 ± 0.616 & 88.058 ± 5.846 \\
0.6 & 92.629 ± 4.660 & 92.268 ± 4.704 & 91.790 ± 4.777 & 90.437 ± 4.619 & 78.279 ± 5.683 & 70.293 ± 4.840 & 89.736 ± 5.886 \\
0.8 & 93.227 ± 4.407 & 92.929 ± 4.537 & 92.682 ± 4.575 & 91.825 ± 4.584 & 83.577 ± 5.449 & 77.870 ± 5.680 & 90.354 ± 5.748 \\
1.0 & 93.381 ± 4.478 & 92.865 ± 4.579 & 92.837 ± 4.656 & 92.354 ± 4.416 & 85.311 ± 5.239 & 80.514 ± 5.594 & 90.369 ± 5.799 \\
1.2 & 93.608 ± 4.329 & 93.374 ± 4.445 & 93.339 ± 4.390 & 92.699 ± 4.447 & 89.673 ± 4.926 & 87.363 ± 5.051 & 89.845 ± 6.026 \\
2.0 & 93.522 ± 4.419 & 93.408 ± 4.362 & 93.346 ± 4.447 & 93.222 ± 4.362 & 91.798 ± 4.772 & 90.671 ± 4.840 & 88.915 ± 6.119 \\
3.0 & 93.445 ± 4.414 & 93.332 ± 4.374 & 93.191 ± 4.356 & 93.127 ± 4.388 & 92.531 ± 4.538 & 92.169 ± 4.463 & 88.321 ± 6.562 \\
4.0 & 93.496 ± 4.366 & 93.365 ± 4.392 & 93.375 ± 4.313 & 93.215 ± 4.309 & 93.065 ± 4.282 & 92.723 ± 4.303 & 87.866 ± 6.702 \\
5.0 & 93.462 ± 4.390 & 93.304 ± 4.448 & 93.127 ± 4.415 & 93.111 ± 4.468 & 93.077 ± 4.445 & 92.750 ± 4.578 & 87.597 ± 6.552 \\
\bottomrule
\end{tabular}
}
\end{table}

\begin{table}[ht]
\centering
\caption{Results for the Liver Disorders dataset.}
{\fontsize{5pt}{7pt}\selectfont
\begin{tabular}{p{0.18cm}|p{1.5153cm}p{1.5153cm}p{1.5153cm}p{1.5153cm}p{1.5153cm}p{1.5153cm}p{1.5153cm}}
\toprule
 BR &     RF(nt\_500) &     RF(qs\_ent) &       RF(mn\_4) &       RF(mn\_8) &       RF(ml\_4) &       RF(ml\_5) &     RF(nf\_all) \\
\midrule
0.2 & 59.507 ± 2.770 & 59.089 ± 2.991 & 59.164 ± 2.827 & 59.485 ± 3.014 & 59.714 ± 2.951 & 59.711 ± 2.870 & 59.213 ± 2.791 \\
0.4 & 58.570 ± 2.799 & 58.240 ± 2.868 & 58.406 ± 2.697 & 58.757 ± 2.760 & 59.109 ± 2.816 & 59.209 ± 2.880 & 58.219 ± 2.778 \\
0.6 & 57.910 ± 2.817 & 57.616 ± 2.845 & 57.923 ± 3.027 & 58.482 ± 2.786 & 58.921 ± 2.846 & 59.206 ± 2.851 & 57.605 ± 2.928 \\
0.8 & 57.391 ± 2.867 & 57.352 ± 2.883 & 57.752 ± 2.896 & 58.259 ± 2.828 & 58.825 ± 2.831 & 59.186 ± 2.841 & 56.903 ± 2.998 \\
1.0 & 57.016 ± 2.746 & 57.061 ± 3.103 & 57.387 ± 2.934 & 57.916 ± 2.934 & 58.635 ± 2.860 & 58.991 ± 2.877 & 56.614 ± 2.884 \\
1.2 & 56.752 ± 2.880 & 56.638 ± 2.901 & 56.961 ± 2.938 & 57.399 ± 2.883 & 58.183 ± 2.858 & 58.446 ± 2.739 & 56.409 ± 3.139 \\
2.0 & 56.160 ± 2.936 & 55.975 ± 3.188 & 56.290 ± 3.053 & 56.956 ± 2.935 & 57.720 ± 2.865 & 57.935 ± 2.867 & 55.569 ± 3.149 \\
3.0 & 55.751 ± 3.008 & 55.612 ± 3.181 & 55.826 ± 3.100 & 56.144 ± 3.027 & 56.830 ± 2.886 & 57.326 ± 3.121 & 55.075 ± 3.227 \\
4.0 & 55.755 ± 3.067 & 55.578 ± 3.134 & 55.721 ± 3.039 & 55.867 ± 3.113 & 56.548 ± 3.034 & 56.877 ± 3.091 & 54.922 ± 3.159 \\
5.0 & 55.645 ± 3.090 & 55.452 ± 3.043 & 55.755 ± 3.006 & 55.776 ± 3.166 & 56.301 ± 3.055 & 56.460 ± 2.977 & 54.812 ± 3.189 \\
\bottomrule
\end{tabular}
}
\end{table}

\begin{table}[ht]
\centering
\caption{Results for the Optical Recognition (Digits) dataset.}
{\fontsize{5pt}{7pt}\selectfont
\begin{tabular}{p{0.18cm}|p{1.5153cm}p{1.5153cm}p{1.5153cm}p{1.5153cm}p{1.5153cm}p{1.5153cm}p{1.5153cm}}
\toprule
 BR &     RF(nt\_500) &     RF(qs\_ent) &       RF(mn\_4) &       RF(mn\_8) &       RF(ml\_4) &       RF(ml\_5) &     RF(nf\_all) \\
\midrule
0.2 & 95.549 ± 0.677 & 95.168 ± 0.742 & 94.761 ± 0.739 & 94.161 ± 0.814 & 93.207 ± 0.878 & 92.702 ± 0.884 & 93.174 ± 0.985 \\
0.4 & 96.386 ± 0.593 & 96.156 ± 0.642 & 95.861 ± 0.679 & 95.341 ± 0.759 & 94.507 ± 0.799 & 94.029 ± 0.788 & 93.866 ± 0.975 \\
0.6 & 96.744 ± 0.572 & 96.553 ± 0.617 & 96.310 ± 0.623 & 95.873 ± 0.664 & 95.094 ± 0.740 & 94.691 ± 0.747 & 94.074 ± 0.977 \\
0.8 & 96.964 ± 0.562 & 96.811 ± 0.544 & 96.581 ± 0.571 & 96.188 ± 0.627 & 95.488 ± 0.713 & 95.104 ± 0.759 & 94.094 ± 1.018 \\
1.0 & 97.108 ± 0.548 & 96.933 ± 0.539 & 96.741 ± 0.580 & 96.378 ± 0.633 & 95.717 ± 0.713 & 95.387 ± 0.711 & 94.014 ± 1.045 \\
1.2 & 97.176 ± 0.535 & 96.986 ± 0.512 & 96.883 ± 0.572 & 96.624 ± 0.598 & 96.108 ± 0.634 & 95.849 ± 0.684 & 93.943 ± 1.061 \\
2.0 & 97.340 ± 0.524 & 97.194 ± 0.499 & 97.085 ± 0.564 & 96.898 ± 0.584 & 96.536 ± 0.598 & 96.267 ± 0.642 & 93.335 ± 1.209 \\
3.0 & 97.396 ± 0.505 & 97.246 ± 0.531 & 97.208 ± 0.527 & 97.096 ± 0.555 & 96.794 ± 0.596 & 96.654 ± 0.601 & 92.561 ± 1.423 \\
4.0 & 97.413 ± 0.519 & 97.255 ± 0.514 & 97.218 ± 0.540 & 97.153 ± 0.556 & 96.977 ± 0.591 & 96.858 ± 0.599 & 91.948 ± 1.544 \\
5.0 & 97.401 ± 0.514 & 97.263 ± 0.511 & 97.233 ± 0.539 & 97.192 ± 0.530 & 97.050 ± 0.541 & 96.944 ± 0.581 & 91.474 ± 1.645 \\
\bottomrule
\end{tabular}
}
\end{table}

\begin{table}[ht]
\centering
\caption{Results for the Parkinsons dataset.}
{\fontsize{5pt}{7pt}\selectfont
\begin{tabular}{p{0.18cm}|p{1.5153cm}p{1.5153cm}p{1.5153cm}p{1.5153cm}p{1.5153cm}p{1.5153cm}p{1.5153cm}}
\toprule
 BR &     RF(nt\_500) &     RF(qs\_ent) &       RF(mn\_4) &       RF(mn\_8) &       RF(ml\_4) &       RF(ml\_5) &     RF(nf\_all) \\
\midrule
0.2 & 84.367 ± 3.323 & 84.487 ± 3.351 & 83.902 ± 3.371 & 82.746 ± 3.191 & 81.979 ± 3.268 & 81.174 ± 3.147 & 84.482 ± 3.362 \\
0.4 & 86.205 ± 3.474 & 86.203 ± 3.337 & 85.618 ± 3.554 & 84.339 ± 3.385 & 84.187 ± 3.226 & 83.464 ± 3.141 & 85.950 ± 3.506 \\
0.6 & 87.223 ± 3.458 & 87.413 ± 3.426 & 86.705 ± 3.585 & 85.462 ± 3.544 & 84.993 ± 3.256 & 84.446 ± 3.294 & 86.554 ± 3.622 \\
0.8 & 87.949 ± 3.369 & 88.085 ± 3.256 & 87.408 ± 3.375 & 86.336 ± 3.557 & 85.757 ± 3.364 & 85.097 ± 3.383 & 86.828 ± 3.765 \\
1.0 & 88.297 ± 3.317 & 88.354 ± 3.123 & 87.874 ± 3.369 & 86.674 ± 3.516 & 86.275 ± 3.346 & 85.687 ± 3.400 & 87.151 ± 3.840 \\
1.2 & 88.536 ± 3.299 & 88.736 ± 3.191 & 88.426 ± 3.260 & 87.629 ± 3.330 & 87.069 ± 3.297 & 86.528 ± 3.318 & 87.195 ± 3.914 \\
2.0 & 89.093 ± 3.270 & 89.150 ± 3.270 & 88.854 ± 3.136 & 88.462 ± 3.320 & 87.782 ± 3.336 & 87.418 ± 3.404 & 87.335 ± 4.205 \\
3.0 & 89.213 ± 3.262 & 89.244 ± 3.162 & 89.139 ± 3.216 & 88.877 ± 3.245 & 88.482 ± 3.285 & 88.215 ± 3.268 & 87.167 ± 4.156 \\
4.0 & 89.306 ± 3.254 & 89.206 ± 3.208 & 89.154 ± 3.123 & 89.031 ± 3.269 & 88.628 ± 3.298 & 88.508 ± 3.275 & 87.000 ± 4.281 \\
5.0 & 89.306 ± 3.234 & 89.160 ± 3.296 & 89.213 ± 3.070 & 89.133 ± 3.130 & 88.810 ± 3.169 & 88.710 ± 3.194 & 86.941 ± 4.166 \\
\bottomrule
\end{tabular}
}
\end{table}

\begin{table}[ht]
\centering
\caption{Results for the Pima Indians Diabetes dataset.}
{\fontsize{5pt}{7pt}\selectfont
\begin{tabular}{p{0.18cm}|p{1.5153cm}p{1.5153cm}p{1.5153cm}p{1.5153cm}p{1.5153cm}p{1.5153cm}p{1.5153cm}}
\toprule
 BR &     RF(nt\_500) &     RF(qs\_ent) &       RF(mn\_4) &       RF(mn\_8) &       RF(ml\_4) &       RF(ml\_5) &     RF(nf\_all) \\
\midrule
0.2 & 76.344 ± 1.557 & 75.767 ± 1.632 & 76.092 ± 1.559 & 76.144 ± 1.580 & 75.973 ± 1.505 & 75.805 ± 1.522 & 76.333 ± 1.659 \\
0.4 & 76.285 ± 1.662 & 76.020 ± 1.555 & 76.152 ± 1.575 & 76.086 ± 1.588 & 76.061 ± 1.561 & 76.040 ± 1.543 & 75.995 ± 1.631 \\
0.6 & 76.161 ± 1.619 & 76.022 ± 1.647 & 76.069 ± 1.552 & 76.099 ± 1.592 & 76.101 ± 1.578 & 75.965 ± 1.555 & 75.912 ± 1.636 \\
0.8 & 76.071 ± 1.633 & 75.937 ± 1.609 & 75.998 ± 1.596 & 76.100 ± 1.711 & 76.058 ± 1.620 & 76.048 ± 1.564 & 75.566 ± 1.743 \\
1.0 & 75.950 ± 1.658 & 75.752 ± 1.627 & 75.921 ± 1.681 & 75.984 ± 1.690 & 76.061 ± 1.636 & 76.148 ± 1.607 & 75.316 ± 1.707 \\
1.2 & 75.883 ± 1.626 & 75.745 ± 1.648 & 75.764 ± 1.705 & 75.999 ± 1.667 & 76.096 ± 1.649 & 76.075 ± 1.616 & 75.107 ± 1.727 \\
2.0 & 75.736 ± 1.648 & 75.579 ± 1.617 & 75.549 ± 1.693 & 75.736 ± 1.733 & 75.937 ± 1.637 & 75.987 ± 1.660 & 74.475 ± 1.832 \\
3.0 & 75.500 ± 1.642 & 75.386 ± 1.675 & 75.349 ± 1.728 & 75.578 ± 1.731 & 75.821 ± 1.641 & 75.811 ± 1.680 & 74.051 ± 1.878 \\
4.0 & 75.517 ± 1.684 & 75.370 ± 1.714 & 75.281 ± 1.667 & 75.446 ± 1.720 & 75.598 ± 1.613 & 75.739 ± 1.636 & 73.600 ± 1.846 \\
5.0 & 75.449 ± 1.700 & 75.231 ± 1.705 & 75.320 ± 1.663 & 75.378 ± 1.669 & 75.575 ± 1.635 & 75.602 ± 1.705 & 73.443 ± 1.910 \\
\bottomrule
\end{tabular}
}
\end{table}

\begin{table}[ht]
\centering
\caption{Results for the Sonar, Mines vs. Rocks dataset.}
{\fontsize{5pt}{7pt}\selectfont
\begin{tabular}{p{0.18cm}|p{1.5153cm}p{1.5153cm}p{1.5153cm}p{1.5153cm}p{1.5153cm}p{1.5153cm}p{1.5153cm}}
\toprule
 BR &     RF(nt\_500) &     RF(qs\_ent) &       RF(mn\_4) &       RF(mn\_8) &       RF(ml\_4) &       RF(ml\_5) &     RF(nf\_all) \\
\midrule
0.2 & 77.678 ± 3.731 & 76.805 ± 3.706 & 76.656 ± 3.867 & 76.327 ± 3.705 & 76.135 ± 3.647 & 75.899 ± 3.627 & 75.820 ± 3.805 \\
0.4 & 78.594 ± 3.922 & 78.212 ± 3.900 & 77.959 ± 3.892 & 77.404 ± 3.868 & 77.752 ± 3.845 & 77.351 ± 3.948 & 76.356 ± 3.951 \\
0.6 & 79.120 ± 3.966 & 79.353 ± 4.033 & 78.697 ± 4.141 & 78.159 ± 3.992 & 78.433 ± 4.140 & 78.099 ± 4.079 & 76.887 ± 4.058 \\
0.8 & 79.572 ± 3.961 & 79.870 ± 3.937 & 79.209 ± 4.078 & 78.442 ± 4.099 & 78.623 ± 4.087 & 78.418 ± 4.111 & 76.803 ± 4.042 \\
1.0 & 80.214 ± 3.992 & 80.240 ± 3.903 & 79.625 ± 4.093 & 78.894 ± 4.004 & 79.130 ± 4.014 & 78.808 ± 4.019 & 76.897 ± 4.227 \\
1.2 & 80.618 ± 4.143 & 80.053 ± 3.981 & 79.750 ± 4.055 & 79.296 ± 4.040 & 79.397 ± 4.041 & 79.055 ± 3.962 & 76.894 ± 4.261 \\
2.0 & 81.236 ± 4.060 & 81.094 ± 3.975 & 80.579 ± 4.141 & 80.276 ± 4.192 & 80.091 ± 4.045 & 79.817 ± 4.206 & 76.548 ± 4.440 \\
3.0 & 81.486 ± 4.057 & 81.308 ± 4.024 & 80.974 ± 3.884 & 80.656 ± 3.927 & 80.514 ± 4.021 & 80.341 ± 4.103 & 75.822 ± 4.701 \\
4.0 & 81.565 ± 4.072 & 81.627 ± 4.007 & 81.317 ± 4.061 & 81.113 ± 4.037 & 80.962 ± 3.934 & 80.668 ± 3.834 & 75.464 ± 4.878 \\
5.0 & 81.548 ± 4.161 & 81.538 ± 3.989 & 81.012 ± 4.123 & 81.161 ± 3.924 & 80.930 ± 3.939 & 80.913 ± 4.007 & 74.950 ± 4.863 \\
\bottomrule
\end{tabular}
}
\end{table}

\begin{table}[ht]
\centering
\caption{Results for the Soybean (Large) dataset.}
{\fontsize{5pt}{7pt}\selectfont
\begin{tabular}{p{0.18cm}|p{1.5153cm}p{1.5153cm}p{1.5153cm}p{1.5153cm}p{1.5153cm}p{1.5153cm}p{1.5153cm}}
\toprule
 BR &     RF(nt\_500) &     RF(qs\_ent) &       RF(mn\_4) &       RF(mn\_8) &       RF(ml\_4) &       RF(ml\_5) &     RF(nf\_all) \\
\midrule
0.2 & 90.450 ± 1.379 & 89.572 ± 1.557 & 89.425 ± 1.555 & 86.323 ± 1.798 & 74.294 ± 2.680 & 67.276 ± 2.540 & 89.222 ± 1.595 \\
0.4 & 91.692 ± 1.346 & 91.319 ± 1.321 & 91.410 ± 1.349 & 90.479 ± 1.479 & 85.207 ± 1.979 & 82.423 ± 2.141 & 91.089 ± 1.541 \\
0.6 & 92.186 ± 1.312 & 91.806 ± 1.382 & 92.050 ± 1.336 & 91.511 ± 1.395 & 87.737 ± 1.699 & 85.760 ± 1.888 & 91.463 ± 1.516 \\
0.8 & 92.386 ± 1.280 & 92.013 ± 1.288 & 92.268 ± 1.291 & 91.938 ± 1.319 & 88.928 ± 1.636 & 87.261 ± 1.727 & 91.628 ± 1.467 \\
1.0 & 92.442 ± 1.282 & 92.101 ± 1.308 & 92.459 ± 1.327 & 92.204 ± 1.316 & 89.641 ± 1.588 & 88.136 ± 1.732 & 91.664 ± 1.474 \\
1.2 & 92.521 ± 1.296 & 92.233 ± 1.293 & 92.440 ± 1.294 & 92.479 ± 1.278 & 90.923 ± 1.434 & 90.044 ± 1.516 & 91.635 ± 1.488 \\
2.0 & 92.606 ± 1.299 & 92.348 ± 1.254 & 92.598 ± 1.304 & 92.695 ± 1.263 & 91.704 ± 1.355 & 91.182 ± 1.450 & 91.505 ± 1.509 \\
3.0 & 92.641 ± 1.300 & 92.412 ± 1.296 & 92.618 ± 1.292 & 92.691 ± 1.240 & 92.153 ± 1.340 & 91.985 ± 1.351 & 91.274 ± 1.564 \\
4.0 & 92.640 ± 1.251 & 92.480 ± 1.278 & 92.561 ± 1.297 & 92.712 ± 1.276 & 92.403 ± 1.278 & 92.220 ± 1.327 & 91.133 ± 1.613 \\
5.0 & 92.662 ± 1.246 & 92.537 ± 1.288 & 92.566 ± 1.290 & 92.681 ± 1.314 & 92.526 ± 1.291 & 92.387 ± 1.289 & 91.027 ± 1.623 \\
\bottomrule
\end{tabular}
}
\end{table}

\begin{table}[ht]
\centering
\caption{Results for the Tic-Tac-Toe Endgame dataset.}
{\fontsize{5pt}{7pt}\selectfont
\begin{tabular}{p{0.18cm}|p{1.5153cm}p{1.5153cm}p{1.5153cm}p{1.5153cm}p{1.5153cm}p{1.5153cm}p{1.5153cm}}
\toprule
 BR &     RF(nt\_500) &     RF(qs\_ent) &       RF(mn\_4) &       RF(mn\_8) &       RF(ml\_4) &       RF(ml\_5) &     RF(nf\_all) \\
\midrule
0.2 & 84.737 ± 1.551 & 84.186 ± 1.683 & 82.142 ± 1.618 & 78.510 ± 1.538 & 76.485 ± 1.409 & 75.429 ± 1.290 & 88.951 ± 2.208 \\
0.4 & 90.749 ± 1.738 & 90.185 ± 1.796 & 88.007 ± 1.783 & 83.849 ± 1.626 & 81.011 ± 1.574 & 79.460 ± 1.552 & 95.987 ± 1.283 \\
0.6 & 93.423 ± 1.661 & 92.680 ± 1.595 & 90.843 ± 1.747 & 86.760 ± 1.740 & 83.504 ± 1.666 & 81.897 ± 1.641 & 96.907 ± 1.053 \\
0.8 & 94.821 ± 1.502 & 93.998 ± 1.537 & 92.594 ± 1.689 & 88.705 ± 1.812 & 85.392 ± 1.809 & 83.569 ± 1.730 & 97.150 ± 0.981 \\
1.0 & 95.590 ± 1.387 & 95.009 ± 1.396 & 93.755 ± 1.592 & 90.200 ± 1.807 & 86.947 ± 1.827 & 85.031 ± 1.774 & 97.131 ± 1.043 \\
1.2 & 96.101 ± 1.310 & 95.585 ± 1.366 & 94.832 ± 1.546 & 92.697 ± 1.652 & 89.772 ± 1.873 & 88.176 ± 1.851 & 97.110 ± 1.027 \\
2.0 & 96.866 ± 1.174 & 96.424 ± 1.192 & 96.055 ± 1.252 & 94.711 ± 1.475 & 92.581 ± 1.781 & 91.113 ± 1.852 & 96.676 ± 1.229 \\
3.0 & 97.162 ± 1.094 & 96.697 ± 1.118 & 96.537 ± 1.200 & 95.793 ± 1.317 & 94.310 ± 1.567 & 93.478 ± 1.713 & 96.235 ± 1.312 \\
4.0 & 97.184 ± 1.063 & 96.819 ± 1.113 & 96.778 ± 1.088 & 96.333 ± 1.246 & 95.201 ± 1.467 & 94.563 ± 1.573 & 95.908 ± 1.438 \\
5.0 & 97.264 ± 1.065 & 96.858 ± 1.128 & 96.873 ± 1.104 & 96.545 ± 1.172 & 95.831 ± 1.306 & 95.226 ± 1.429 & 95.608 ± 1.482 \\
\bottomrule
\end{tabular}
}
\end{table}

\begin{table}[ht]
\centering
\caption{Results for the Thyroid Disease dataset.}
{\fontsize{5pt}{7pt}\selectfont
\begin{tabular}{p{0.18cm}|p{1.5153cm}p{1.5153cm}p{1.5153cm}p{1.5153cm}p{1.5153cm}p{1.5153cm}p{1.5153cm}}
\toprule
 BR &     RF(nt\_500) &     RF(qs\_ent) &       RF(mn\_4) &       RF(mn\_8) &       RF(ml\_4) &       RF(ml\_5) &     RF(nf\_all) \\
\midrule
0.2 & 92.821 ± 2.404 & 92.992 ± 2.334 & 93.078 ± 2.393 & 92.924 ± 2.317 & 89.785 ± 2.068 & 82.280 ± 3.670 & 93.146 ± 2.386 \\
0.4 & 95.148 ± 1.951 & 94.949 ± 1.940 & 95.049 ± 2.068 & 94.822 ± 2.099 & 93.696 ± 2.220 & 93.371 ± 2.124 & 93.927 ± 2.296 \\
0.6 & 95.529 ± 1.952 & 95.478 ± 1.831 & 95.384 ± 2.054 & 95.166 ± 2.103 & 94.431 ± 2.199 & 94.229 ± 2.256 & 94.203 ± 2.366 \\
0.8 & 95.552 ± 1.945 & 95.675 ± 1.848 & 95.366 ± 1.938 & 95.087 ± 2.067 & 94.631 ± 2.131 & 94.496 ± 2.179 & 94.142 ± 2.378 \\
1.0 & 95.542 ± 1.926 & 95.766 ± 1.841 & 95.331 ± 1.981 & 95.070 ± 2.055 & 94.710 ± 2.176 & 94.531 ± 2.210 & 94.080 ± 2.362 \\
1.2 & 95.624 ± 1.903 & 95.840 ± 1.777 & 95.482 ± 1.917 & 95.240 ± 1.968 & 95.010 ± 2.091 & 94.870 ± 2.109 & 94.078 ± 2.288 \\
2.0 & 95.459 ± 1.935 & 95.635 ± 1.858 & 95.328 ± 1.924 & 95.112 ± 2.057 & 94.964 ± 2.034 & 94.815 ± 2.110 & 93.763 ± 2.308 \\
3.0 & 95.321 ± 2.021 & 95.565 ± 1.853 & 95.152 ± 2.030 & 95.029 ± 2.010 & 94.882 ± 2.139 & 94.778 ± 2.179 & 93.689 ± 2.327 \\
4.0 & 95.217 ± 2.047 & 95.523 ± 1.882 & 95.042 ± 2.059 & 94.924 ± 2.080 & 94.847 ± 2.193 & 94.768 ± 2.280 & 93.622 ± 2.378 \\
5.0 & 95.163 ± 2.044 & 95.423 ± 1.924 & 94.977 ± 2.058 & 94.898 ± 2.041 & 94.858 ± 2.108 & 94.724 ± 2.113 & 93.601 ± 2.347 \\
\bottomrule
\end{tabular}
}
\end{table}

\begin{table}[ht]
\centering
\caption{Results for the Vehicle Silhouettes dataset.}
{\fontsize{5pt}{7pt}\selectfont
\begin{tabular}{p{0.18cm}|p{1.5153cm}p{1.5153cm}p{1.5153cm}p{1.5153cm}p{1.5153cm}p{1.5153cm}p{1.5153cm}}
\toprule
 BR &     RF(nt\_500) &     RF(qs\_ent) &       RF(mn\_4) &       RF(mn\_8) &       RF(ml\_4) &       RF(ml\_5) &     RF(nf\_all) \\
\midrule
0.2 & 72.444 ± 1.686 & 72.441 ± 1.739 & 72.170 ± 1.746 & 71.441 ± 1.810 & 70.228 ± 1.823 & 69.615 ± 1.790 & 72.351 ± 1.817 \\
0.4 & 73.467 ± 1.615 & 73.362 ± 1.587 & 73.238 ± 1.683 & 72.954 ± 1.710 & 71.817 ± 1.768 & 71.322 ± 1.774 & 73.385 ± 1.837 \\
0.6 & 73.932 ± 1.618 & 73.790 ± 1.564 & 73.845 ± 1.625 & 73.434 ± 1.670 & 72.490 ± 1.695 & 71.973 ± 1.794 & 73.595 ± 1.680 \\
0.8 & 74.286 ± 1.606 & 74.070 ± 1.560 & 74.118 ± 1.673 & 73.737 ± 1.597 & 72.794 ± 1.692 & 72.355 ± 1.770 & 73.597 ± 1.682 \\
1.0 & 74.397 ± 1.591 & 74.227 ± 1.509 & 74.303 ± 1.614 & 73.998 ± 1.663 & 73.005 ± 1.694 & 72.568 ± 1.683 & 73.656 ± 1.709 \\
1.2 & 74.442 ± 1.581 & 74.322 ± 1.568 & 74.347 ± 1.608 & 74.212 ± 1.590 & 73.451 ± 1.695 & 73.160 ± 1.646 & 73.471 ± 1.747 \\
2.0 & 74.557 ± 1.519 & 74.487 ± 1.561 & 74.566 ± 1.615 & 74.422 ± 1.555 & 73.948 ± 1.619 & 73.616 ± 1.683 & 73.220 ± 1.765 \\
3.0 & 74.506 ± 1.542 & 74.434 ± 1.576 & 74.531 ± 1.562 & 74.455 ± 1.597 & 74.127 ± 1.618 & 74.071 ± 1.574 & 72.681 ± 1.856 \\
4.0 & 74.544 ± 1.480 & 74.547 ± 1.582 & 74.480 ± 1.529 & 74.522 ± 1.562 & 74.214 ± 1.536 & 74.138 ± 1.563 & 72.343 ± 1.885 \\
5.0 & 74.583 ± 1.512 & 74.442 ± 1.576 & 74.544 ± 1.577 & 74.505 ± 1.549 & 74.326 ± 1.665 & 74.276 ± 1.495 & 72.091 ± 1.918 \\
\bottomrule
\end{tabular}
}
\end{table}

\begin{table}[ht]
\centering
\caption{Results for the Vowel Recognition dataset.}
{\fontsize{5pt}{7pt}\selectfont
\begin{tabular}{p{0.18cm}|p{1.5153cm}p{1.5153cm}p{1.5153cm}p{1.5153cm}p{1.5153cm}p{1.5153cm}p{1.5153cm}}
\toprule
 BR &     RF(nt\_500) &     RF(qs\_ent) &       RF(mn\_4) &       RF(mn\_8) &       RF(ml\_4) &       RF(ml\_5) &     RF(nf\_all) \\
\midrule
0.2 & 84.796 ± 2.104 & 81.707 ± 2.151 & 80.631 ± 2.235 & 76.111 ± 2.369 & 73.027 ± 2.420 & 70.847 ± 2.581 & 77.066 ± 2.491 \\
0.4 & 88.552 ± 1.902 & 86.939 ± 2.015 & 85.794 ± 2.090 & 82.279 ± 2.149 & 78.848 ± 2.342 & 76.617 ± 2.353 & 82.283 ± 2.336 \\
0.6 & 90.170 ± 1.852 & 89.077 ± 1.852 & 88.036 ± 1.943 & 85.067 ± 2.124 & 81.585 ± 2.264 & 79.443 ± 2.321 & 84.474 ± 2.194 \\
0.8 & 90.981 ± 1.737 & 90.127 ± 1.732 & 89.185 ± 1.881 & 86.562 ± 2.053 & 83.281 ± 2.128 & 81.293 ± 2.266 & 85.400 ± 2.053 \\
1.0 & 91.415 ± 1.745 & 90.697 ± 1.694 & 89.849 ± 1.804 & 87.463 ± 2.015 & 84.339 ± 2.210 & 82.412 ± 2.172 & 85.844 ± 2.090 \\
1.2 & 91.657 ± 1.692 & 91.089 ± 1.653 & 90.576 ± 1.767 & 89.089 ± 1.895 & 86.603 ± 2.078 & 85.111 ± 2.154 & 86.088 ± 2.145 \\
2.0 & 92.146 ± 1.616 & 91.751 ± 1.686 & 91.428 ± 1.663 & 90.328 ± 1.796 & 88.485 ± 1.956 & 87.211 ± 2.098 & 85.754 ± 2.149 \\
3.0 & 92.285 ± 1.651 & 91.964 ± 1.601 & 91.718 ± 1.651 & 91.034 ± 1.757 & 89.755 ± 1.855 & 88.989 ± 1.876 & 84.890 ± 2.276 \\
4.0 & 92.278 ± 1.662 & 91.929 ± 1.679 & 91.785 ± 1.722 & 91.363 ± 1.701 & 90.404 ± 1.804 & 89.876 ± 1.829 & 84.246 ± 2.288 \\
5.0 & 92.182 ± 1.667 & 91.922 ± 1.652 & 91.730 ± 1.719 & 91.492 ± 1.683 & 90.832 ± 1.729 & 90.358 ± 1.742 & 83.619 ± 2.390 \\
\bottomrule
\end{tabular}
}
\end{table}

\begin{table}[ht]
\centering
\caption{Results for the Wine dataset.}
{\fontsize{5pt}{7pt}\selectfont
\begin{tabular}{p{0.18cm}|p{1.5153cm}p{1.5153cm}p{1.5153cm}p{1.5153cm}p{1.5153cm}p{1.5153cm}p{1.5153cm}}
\toprule
 BR &     RF(nt\_500) &     RF(qs\_ent) &       RF(mn\_4) &       RF(mn\_8) &       RF(ml\_4) &       RF(ml\_5) &     RF(nf\_all) \\
\midrule
0.2 & 97.548 ± 1.432 & 96.649 ± 1.856 & 96.674 ± 1.847 & 96.444 ± 1.891 & 96.413 ± 1.915 & 96.435 ± 1.965 & 95.716 ± 2.252 \\
0.4 & 97.632 ± 1.363 & 97.067 ± 1.751 & 97.034 ± 1.743 & 96.840 ± 1.767 & 96.489 ± 1.785 & 96.447 ± 1.870 & 95.309 ± 2.617 \\
0.6 & 97.694 ± 1.358 & 97.346 ± 1.619 & 97.298 ± 1.603 & 97.191 ± 1.642 & 96.677 ± 1.761 & 96.581 ± 1.755 & 95.073 ± 2.973 \\
0.8 & 97.728 ± 1.347 & 97.368 ± 1.575 & 97.424 ± 1.519 & 97.331 ± 1.563 & 96.691 ± 1.697 & 96.579 ± 1.758 & 95.132 ± 3.017 \\
1.0 & 97.764 ± 1.381 & 97.396 ± 1.578 & 97.455 ± 1.534 & 97.357 ± 1.560 & 96.728 ± 1.773 & 96.581 ± 1.807 & 94.927 ± 3.137 \\
1.2 & 97.809 ± 1.312 & 97.416 ± 1.630 & 97.441 ± 1.528 & 97.385 ± 1.601 & 96.969 ± 1.638 & 96.812 ± 1.721 & 94.916 ± 3.081 \\
2.0 & 97.761 ± 1.340 & 97.416 ± 1.663 & 97.567 ± 1.513 & 97.455 ± 1.559 & 97.039 ± 1.675 & 96.938 ± 1.789 & 94.067 ± 3.291 \\
3.0 & 97.654 ± 1.465 & 97.362 ± 1.615 & 97.489 ± 1.541 & 97.452 ± 1.581 & 97.211 ± 1.682 & 97.051 ± 1.713 & 93.492 ± 3.291 \\
4.0 & 97.649 ± 1.445 & 97.357 ± 1.639 & 97.500 ± 1.509 & 97.511 ± 1.513 & 97.334 ± 1.598 & 97.272 ± 1.595 & 93.101 ± 3.373 \\
5.0 & 97.640 ± 1.421 & 97.329 ± 1.648 & 97.492 ± 1.536 & 97.438 ± 1.560 & 97.289 ± 1.612 & 97.236 ± 1.604 & 92.812 ± 3.454 \\
\bottomrule
\end{tabular}
}
\end{table}

\begin{table}[ht]
\centering
\caption{Results for the Ringnorm dataset.}
{\fontsize{5pt}{7pt}\selectfont
\begin{tabular}{p{0.18cm}|p{1.5153cm}p{1.5153cm}p{1.5153cm}p{1.5153cm}p{1.5153cm}p{1.5153cm}p{1.5153cm}}
\toprule
 BR &     RF(nt\_500) &     RF(qs\_ent) &       RF(mn\_4) &       RF(mn\_8) &       RF(ml\_4) &       RF(ml\_5) &     RF(nf\_all) \\
\midrule
0.2 & 89.752 ± 2.383 & 88.622 ± 2.779 & 89.057 ± 2.718 & 90.107 ± 2.605 & 83.540 ± 3.072 & 81.078 ± 3.104 & 81.605 ± 3.569 \\
0.4 & 92.597 ± 2.220 & 91.437 ± 2.413 & 91.802 ± 2.314 & 92.443 ± 2.212 & 89.123 ± 2.691 & 87.778 ± 2.827 & 86.065 ± 3.641 \\
0.6 & 92.717 ± 2.390 & 91.538 ± 2.526 & 91.962 ± 2.455 & 92.335 ± 2.352 & 89.738 ± 2.701 & 88.995 ± 2.848 & 86.747 ± 3.468 \\
0.8 & 92.370 ± 2.473 & 91.515 ± 2.582 & 92.060 ± 2.533 & 92.192 ± 2.413 & 89.898 ± 2.675 & 89.288 ± 2.742 & 86.725 ± 3.669 \\
1.0 & 92.173 ± 2.457 & 91.325 ± 2.692 & 91.710 ± 2.543 & 91.957 ± 2.376 & 89.828 ± 2.619 & 89.232 ± 2.687 & 86.767 ± 3.679 \\
1.2 & 91.977 ± 2.513 & 91.010 ± 2.693 & 91.407 ± 2.617 & 91.657 ± 2.539 & 90.160 ± 2.638 & 89.562 ± 2.832 & 86.563 ± 3.737 \\
2.0 & 91.322 ± 2.523 & 90.442 ± 2.754 & 91.068 ± 2.593 & 91.213 ± 2.590 & 89.865 ± 2.754 & 89.453 ± 2.837 & 85.672 ± 3.747 \\
3.0 & 90.670 ± 2.640 & 89.985 ± 2.752 & 90.365 ± 2.661 & 90.450 ± 2.597 & 89.757 ± 2.755 & 89.432 ± 2.788 & 84.672 ± 4.002 \\
4.0 & 90.398 ± 2.648 & 89.603 ± 2.710 & 90.092 ± 2.730 & 90.173 ± 2.691 & 89.680 ± 2.698 & 89.405 ± 2.692 & 84.085 ± 3.918 \\
5.0 & 90.173 ± 2.731 & 89.312 ± 2.864 & 89.882 ± 2.666 & 89.953 ± 2.717 & 89.555 ± 2.733 & 89.473 ± 2.724 & 83.438 ± 3.954 \\
\bottomrule
\end{tabular}
}
\end{table}

\begin{table}[ht]
\centering
\caption{Results for the Threenorm dataset.}
{\fontsize{5pt}{7pt}\selectfont
%
}
\end{table}

\begin{table}[ht]
\centering
\caption{Results for the Twonorm dataset.}
{\fontsize{5pt}{7pt}\selectfont
%
}
\end{table}

\begin{table}[ht]
\centering
\caption{Results for the Waveform dataset.}
{\fontsize{5pt}{7pt}\selectfont
%
}
\end{table}

\clearpage
\begin{table}[h]
\centering
\caption{Results for the LED Display Domain dataset.}
{\fontsize{5pt}{7pt}\selectfont
%
}
\end{table}

\clearpage
\section{Bootstrap Rate Curves}
\label{appendix:BR_curves}
\setcounter{figure}{0}
\begin{figure}[h!]
\centering
  \includegraphics[width=\textwidth, height=0.888\textheight, keepaspectratio]{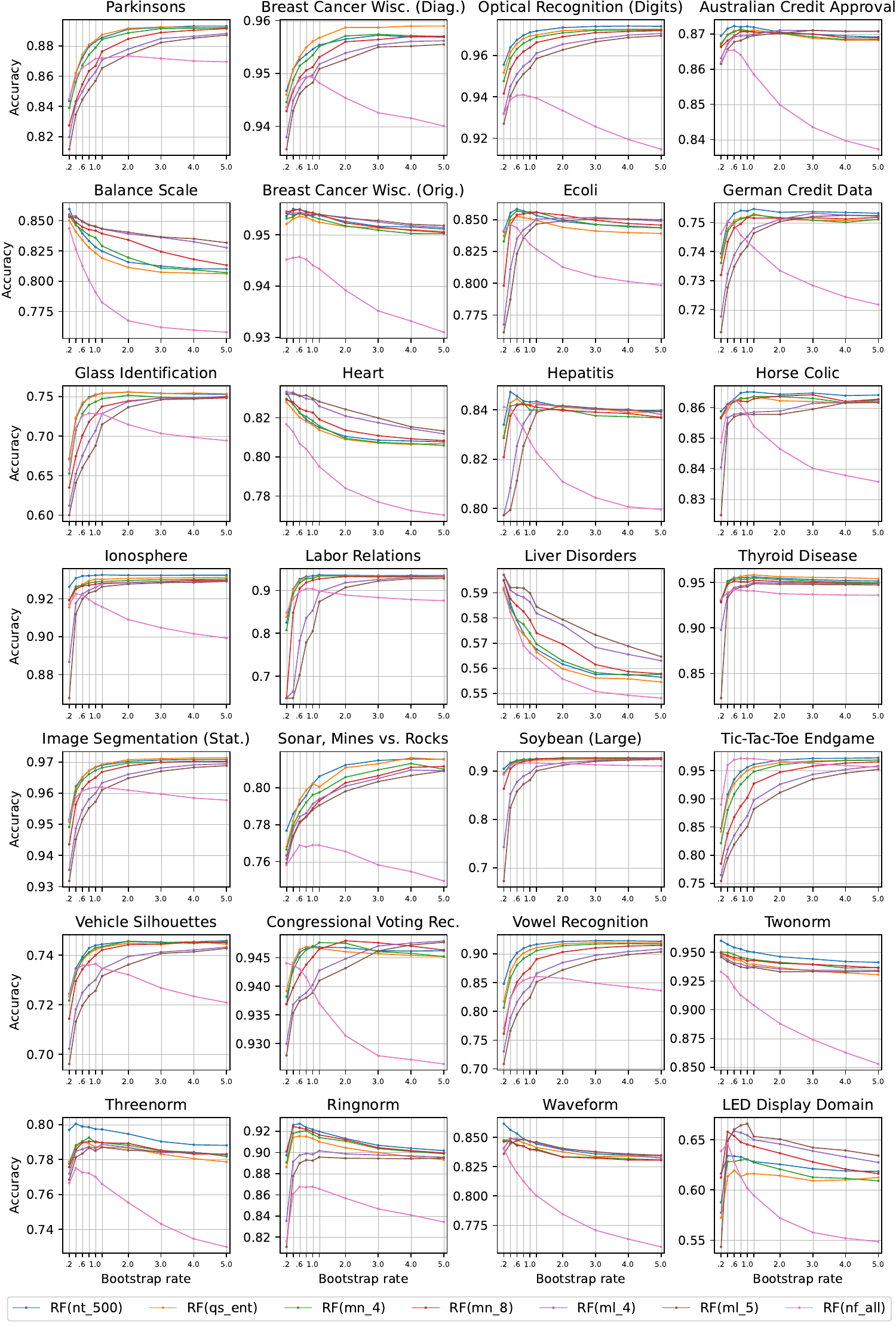}
\caption{Characteristics of BR curves for datasets not shown in Figure~\ref{fig:BR_characteristics_main}.}
\label{fig:BR_characteristics_appendix}  
\end{figure}

\bibliographystyle{elsarticle-num-names} 
\bibliography{references}

\end{document}